\documentclass[journal]{IEEEtran}
\usepackage{amsmath,amsfonts}
\usepackage{amsthm,amssymb}
\usepackage{mathrsfs}
\usepackage{CJKutf8}
\usepackage[hidelinks]{hyperref}
\usepackage{algorithmic}
\usepackage{algorithm}
\usepackage{array}
\usepackage[caption=false,font=normalsize,labelfont=sf,textfont=sf]{subfig}
\usepackage{textcomp}
\usepackage{stfloats}
\usepackage{url}
\usepackage{verbatim}
\usepackage{graphicx}
\usepackage{cite}

\usepackage{soul}
\usepackage{bm}
\usepackage{booktabs}
\usepackage{txfonts}
\usepackage{booktabs}
\usepackage{multirow}
\usepackage{caption3}
\usepackage{xcolor}
\usepackage{pifont}
\usepackage{threeparttable}

\usepackage{hyperref}
\usepackage{float}

\usepackage{textcomp}
\usepackage{stfloats}
\usepackage{url}
\usepackage{verbatim}
\usepackage{graphicx}

\usepackage{cite}

\usepackage{soul}
\usepackage{bm}
\usepackage{booktabs}
\usepackage{txfonts}
\usepackage{booktabs}
\usepackage{multirow}
\usepackage{caption3}
\usepackage{caption}
\usepackage{xcolor}
\usepackage{float}
\usepackage[table]{xcolor} 
\definecolor{darkgray}{rgb}{0.7, 0.7, 0.7} 
\definecolor{gray}{rgb}{0.8, 0.8, 0.8}     
\definecolor{lightgray}{rgb}{0.9, 0.9, 0.9} 

\usepackage{pifont}
\usepackage{threeparttable}

\begin{document}
\begin{CJK*}{UTF8}{gbsn}
\title{Efficient Remote Sensing Instance Segmentation with Linear-Time State Space Distilled Visual Foundation Models}

\author{Qinzhe~Yang$^\dag$, Keyan~Chen$^\dag$,~Jia~Xu,\\Zhenwei~Shi,~\IEEEmembership{Senior~Member,~IEEE},~and~Zhengxia~Zou$^\star$,~\IEEEmembership{Senior~Member,~IEEE}
\thanks{The work was supported by the National Natural Science Foundation of China under Grant 62125102, 62471014, U24B20177, U25A20401, and 623B2013, and the Fundamental Research Funds for the Central Universities. \emph{(Corresponding author: Zhengxia Zou (e-mail: zhengxiazou@buaa.edu.cn))} 
}
\thanks{Qinzhe Yang is with Shen Yuan Honors College, Beihang University, Beijing 100191, China. Keyan Chen, Zhenwei Shi, and Zhengxia Zou are with the Department of Aerospace Intelligent Science and Technology, School of Astronautics, and with the State Key Laboratory of Virtual Reality Technology and Systems, Beihang University, Beijing 100191, China. Jia Xu is with the Qian Xuesen
Laboratory of Space Technology, China Academy of Space Technology,
Beijing 100094, China. $^\dag$ These authors contributed equally to this paper. 

}
}

\maketitle

\begin{abstract}

The computational complexity of Transformers scales quadratically with the number of tokens, which significantly constrains the efficiency of vision models, particularly recent ViT-based foundation models in dense prediction tasks. Instance segmentation, a typical dense visual prediction task in the remote sensing field, faces similar challenges. In this paper, inspired by the recent advances of knowledge distillation in large language models, we introduce RS4D - a new remote sensing instance segmentation method with linear computational complexity, which addresses the inefficiency of long sequence modeling through distilled state space modeling (SSM). We propose an adaptive noise and masking knowledge distillation training method for pre-training lightweight SSM backbones, which effectively compresses knowledge from the vast self-attention space into a compact, dense linear state space. We also design a remote sensing image instance segmentation architecture based on this lightweight visual encoder, where we explore variants of three different backbones and two segmentation heads. Extensive experiments are conducted on multiple benchmark datasets, including SSDD, WHU, and NWPU. Compared to ViT-based approaches, our proposed SSM backbone
achieves an $\times 8$ reduction in parameters and a $\times 9$ reduction in FLOPs while maintaining comparable or superior accuracy to both ViT- and CNN-based instance segmentation methods. The implementation codes have been publicly available at \url{https://github.com/QinzheYang/RS4D}.

\end{abstract}


\section{Introduction}

\IEEEPARstart{T}{he} advancement of deep learning models has accelerated the development of instance segmentation methods \cite{object20,survey_of_remote_sensing,chen2023target,zheng2023farseg++,zhao2023inherit}. Initially, these methods primarily relied on Convolutional Neural Network (CNN) architectures \cite{bolya2019yolact,wang2020solo,xie2020polarmask,sofiiuk2019adaptis, HQ-ISNet,yuan2017learning,huan2021unmixing}. With the emergence of visual foundation models, Transformer-based architectures have gained prominence in image segmentation tasks, exemplified by the Segment Anything Model (SAM) \cite{SAM}. These approaches have been successfully extended to instance segmentation in remote sensing applications \cite{RSPrompter, wang2024samrs}. They often incorporate pre-trained foundation models, such as Vision Transformer (ViT) \cite{attention_transfor}, as their backbone architecture. While these models demonstrate superior accuracy compared to traditional CNN-based approaches, they require significantly more parameters and computational resources. This is because dense prediction typically requires processing high-resolution feature maps and more tokens, so the quadratic self-attention cost in ViTs leads to rapidly increasing computation and memory usage as resolution grows.

\begin{figure}[H]
\centering
\includegraphics[width=1\linewidth]{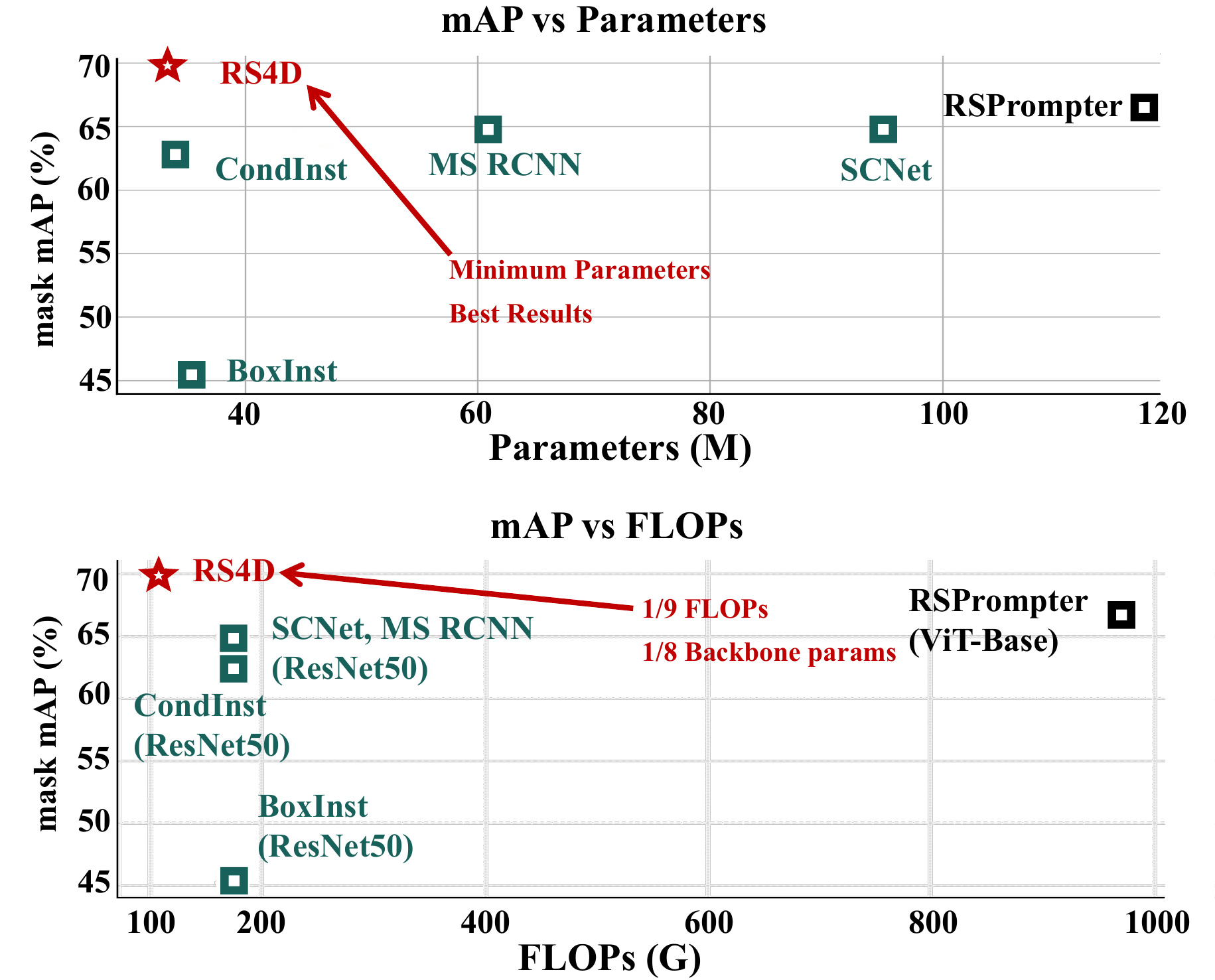}
    \vspace{-4ex}
\caption{
Comparison with state-of-the-art methods on the SSDD dataset evaluating mask mAP, model parameters, and FLOPs.
}
\label{fig:det1}
\end{figure}

\begin{figure*}
\centering
\includegraphics[width=1\linewidth]{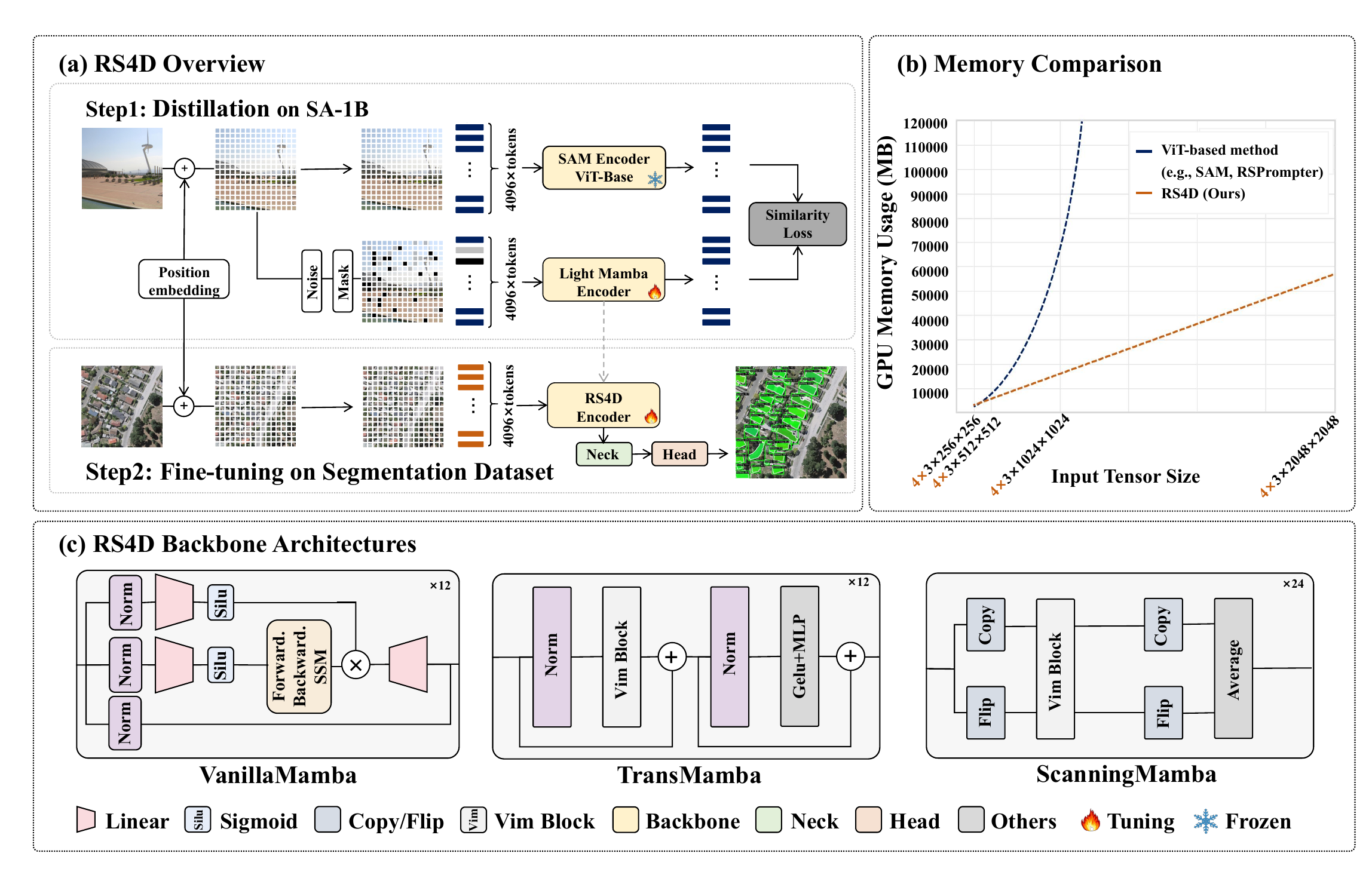}
\caption{
An overview of our proposed RS4D. The training process comprises two stages, as illustrated in panel (a): 1) knowledge distillation from the ViT-based SAM encoder to our lightweight Mamba encoder, using SA-1B dataset\cite{SAM}, and 2) model fine-tuning on domain-specific remote sensing datasets. Panel (b) demonstrates that, in contrast to RSPrompter, our model achieves nearly linear computational complexity and reduces GPU memory consumption by 76\% when processing $1024 \times1024$ pixel images. Panel (c) shows our exploration of multiple model implementations within the proposed framework, featuring three distinct backbone architectures.} 
\label{fig:overall}
\end{figure*}

Remote sensing image instance segmentation fundamentally operates as a dense prediction task, which manifests as a long-sequence prediction problem \cite{RSPrompter,survey_of_remote_sensing, chen2022resolution,hong2024spectralgpt,zhao2023vehicle,chen2025rsrefseg}. For example, processing a $1024 \times 1024$ pixel input image with a spatial resolution of $16 \times 16$ pixels per token generates a sequence length of 4096 tokens. However, the self-attention mechanism in Transformer architectures exhibits quadratic computational complexity with respect to sequence length \cite{flashattention,poolingformer,fastformer,longformer, chen2021building,li2024box2mask,wu2025fully,sun2024learning}. This computational burden leads to substantial memory requirements and slow inference speeds when applying Transformer models to remote sensing instance segmentation tasks. Consequently, these models prove impractical for time-sensitive applications or resource-constrained edge devices, particularly in scenarios such as in-orbit processing.

The State Space Model (SSM) \cite{SSM_in_control,S4,ViM,combine_cnn_with_ssm,hippo} has emerged as a promising framework for processing long sequences, gaining significant attention due to its linear computational complexity. In processing sequential data, the model's state and output equations are interwoven in a way that mathematically unfolds into a structure analogous to one-dimensional convolution operations. This way of mathematical modeling enables efficient computation and makes SSM particularly well-suited for dense prediction tasks, such as instance segmentation. Compared to Transformer-based architectures, SSM offers enhanced performance potential and optimization capabilities. In particular, the linear-time token mixing in SSMs enables global context modeling on long sequences at substantially lower memory/computation than quadratic attention, which can translate into a better accuracy–efficiency trade-off for high-resolution dense prediction when paired with appropriate architectural design.

In SSM instance segmentation, there have already been many outstanding works. MambaInst\cite{wang2025mambainst} achieves efficient instance segmentation by designing the LightSSM Block and FRDown module, maintaining linear computational complexity. Spatial-Mamba\cite{xiao2024spatial} tends to enhance the spatial relationship capture of SSM through convolution. However, this paper aimed at presenting a novel lightweight visual pre-training approach for Vision Transformers utilizing knowledge distillation to compress information from extensive self-attention spaces into compact linear state spaces. We name our method \textbf{RS4D} - \textbf{R}emote \textbf{S}ensing instance \textbf{S}egmentation with linear-time \textbf{S}tate
\textbf{S}pace \textbf{D}istilled visual foundation models. In recent research on large language models, especially in reasoning models, model distillation has proven to be an effective lightweight strategy and has demonstrated significant potential in architectures like Llama, DeepSeek, and others\cite{touvron2023llama,guo2025deepseek}. We develop a new visual encoder network architecture leveraging SSMs that efficiently encodes complex knowledge with minimal parametric overhead. Our framework encompasses multiple architectural variations, including three distinct SSM backbones and two instance segmentation heads, each demonstrating robust performance. Comprehensive evaluations on established benchmark datasets (SSDD \cite{SSDD}, WHU \cite{ji2018fully}, and NWPU \cite{NWPU}) validate our approach, with results presented in Fig. \ref{fig:det1}. The proposed RS4D architecture achieves significant computational efficiency, reducing FLOPs by $\times 9$ compared to ViT-based methods and by $\times 6$ relative to ConvNet approaches, while maintaining comparable accuracy. Most notably, RS4D matches the performance metrics of RSPrompter~\cite{RSPrompter} (with pre-trained ViT-based encoder) while requiring only 1/4 of the model parameters and GPU memory during training operations.

The contributions of this paper can be summarized as follows:
1) We present a novel remote-sensing-oriented instance segmentation architecture optimized for dense prediction, leveraging State Space Models (SSMs). The proposed approach combines an efficient scanning mechanism with knowledge distillation techniques derived from large-scale Vision Transformers, thereby achieving both architectural efficiency and effective compression of foundational model knowledge. 

2) We systematically investigate various architectural configurations, implementing three distinct SSM backbones and two instance segmentation heads. Through this exploration, we identify design structures that are more suitable for remote sensing dense prediction tasks than existing methods.

3) Comprehensive experiments conducted on the SSDD, WHU, and NWPU datasets reveal that our model achieves performance parity or superiority compared to existing CNN- and ViT-based instance segmentation methods, while significantly reducing computational demands. Specifically, the model requires only 1/8 of the parameters and 1/9 of the FLOPs relative to baseline architectures.

\section{Related Works}

This section reviews four key areas relevant to our proposed method: instance segmentation, visual foundation models, lightweight models, and state-space models.

\subsection{Instance Segmentation}

Instance segmentation is a fundamental computer vision task that integrates pixel-level semantic segmentation with instance-level object detection. The methods can be classified into two primary categories: two-stage and one-stage methods. Representative two-stage methods include Mask R-CNN \cite{Mask_r-cnn}, MS R-CNN \cite{Mask_scoring_r-cnn}, HTC \cite{HTC}, and HQ-ISNet \cite{HQ-ISNet}, while one-stage methods include YOLACT \cite{bolya2019yolact}, PolarMask \cite{xie2020polarmask}, CondInst \cite{Coninst}, SOLOv2 \cite{solov2}, and MaskFormer \cite{maskformer}, etc. The difference between the two methods is that the one-stage method directly generates predictions from the image, while the two-stage method first generates candidate regions and then performs classification and regression. Two-stage methods typically follow a proposal-and-refine pipeline, often achieving strong accuracy but with extra cost from proposal and RoI processing. In contrast, one-stage methods predict masks directly from dense features without explicit RoI proposals, usually offering higher efficiency at the cost of more careful design to match two-stage quality.

Recent advances in foundation models have catalyzed significant progress in segmentation tasks. In particular, prompt-based segmentation methods, exemplified by the Segment Anything Model (SAM), have emerged as versatile solutions for diverse segmentation challenges \cite{SAM, efficientsam}. The integration of large language models has opened new avenues in language-guided segmentation research \cite{zhang2025psalm}. Furthermore, researchers have successfully adapted the denoising diffusion process, inherent in diffusion models, to segmentation applications \cite{ji2023ddp,chen2023generative}. Recent investigations have also demonstrated the potential of zero-shot segmentation by leveraging segmentation knowledge extracted from pre-trained foundation models \cite{zhou2022extract,zhou2024image}.

Remote sensing instance segmentation poses additional challenges compared with natural-image benchmarks, including very high spatial resolution, small and sparsely distributed targets, and modality variations\cite{survey_of_remote_sensing}. To address these characteristics, several methods have been proposed specifically for remote sensing imagery, such as HQ-ISNet\cite{HQ-ISNet} and CATNet\cite{CATNet}. In this paper, we include these RS-oriented methods as representative remote-sensing baselines and highlight how our approach differs by transferring foundation-model segmentation priors into a lightweight linear-time SSM backbone.

\subsection{Visual Foundation Models}

SAM exemplifies visual foundation models specialized in image segmentation \cite{SAM}. Its architecture comprises three key components: a pre-trained image encoder, a prompt encoder, and a decoder. The image encoder converts input images into deep feature representations, while the prompt encoder transforms user-defined inputs (target points, masks, and boxes) into embeddings. The decoder then synthesizes outputs from both encoders to generate foreground masks at specified locations. SAM has demonstrated exceptional performance across various general vision applications and exhibits robust generalization capabilities \cite{RSPrompter,autosam}, spurring extensive research and numerous task-specific variants \cite{mobilesamv2,efficientsam,fastersam}.

Building on SAM, several new image segmentation methods have been introduced. RSPrompter \cite{RSPrompter}, which implements an automatic prompter by utilizing intermediate outputs from the image encoder to enable automatic instance segmentation. MobileSAM \cite{mobilesamv2} and EfficientSAM \cite{efficientsam} focus on compressing the image encoder through distillation into a smaller ViT. SlimSAM \cite{slimsam} combines pruning and distillation techniques to develop a lightweight SAM variant. Despite these advancements, these approaches encounter a fundamental limitation: their performance and computational requirements remain intrinsically bound to the transformer architecture. This inherent dependency creates a significant challenge in optimizing the trade-off between processing efficiency and segmentation accuracy, making it difficult to achieve an optimal balance between computational speed and performance quality.

Beyond general-purpose promptable segmentation models such as SAM, recent studies have explored adapting SAM-like priors to remote sensing imagery via prompt generation, domain-aware fine-tuning, and RS-specific prompting strategies. Representative directions include segmentation-guided SAM variants\cite{zhang2024rs} and remote-sensing prompters that generate prompts from RS imagery\cite{RSPrompter}. These works suggest that foundation-model priors can benefit RS segmentation. However, most approaches still rely on computationally heavy backbones. Our work focuses on distilling SAM’s segmentation priors into a compact student backbone with linear-time complexity.

\subsection{Lightweight Models}

The lightweight deep learning methods can be categorized into five main groups \cite{KD,Deep_Compression_pruning,SVD,squeezenet}: knowledge distillation, model pruning, quantization, low-rank decomposition, and lightweight model design. Knowledge distillation transfers computational knowledge from a complex teacher model to a more compact student model \cite{KD,fitnets,attention_transfor,FSP_matrix}. Network pruning enhances model efficiency by eliminating redundant or less significant weights \cite{Deep_Compression_pruning,FSP_matrix,BANs,RKD}, effectively reducing the network's complexity. Model quantization research primarily focuses on two directions: multi-network weight sharing and single-node weight reduction \cite{Deep_Compression_pruning,hash,8-bit_fixed_point,Dynamic_fixed_point}. Low-rank decomposition techniques, such as SVD and PQ, decompose the model's original tensor weights into multiple low-rank tensors \cite{SVD}. Unlike these compression strategies, the lightweight design emphasizes building efficient network architectures from inception, typically incorporating lightweight CNN architectures and optimized neural operators \cite{squeezenet,mobilenets,shufflenet,xception}.

While previous approaches have predominantly centered on convolutional or attention-based network designs, our research introduces a novel perspective by employing a visual state space model. This approach facilitates knowledge transfer from complex, attention-mechanism-based foundation models to a more efficient state space model network.

\subsection{State Space Models}

The State Space Model (SSM) refers to a class of mathematical frameworks for representing discrete dynamical systems that describe both the system's temporal evolution and the relationship between hidden states and observable data. Originally developed in control theory, SSM processes continuous sequences to generate predictive outputs. The model is characterized by two fundamental equations: the state equation, which defines the relationship between inputs and previous states through matrix operations, and the output equation, which determines how the state matrix is transformed into observable outputs:
\begin{equation}
\begin{split}
h_k&=Ah_{k-1}+Bx_k \\
y_k&=Ch_k
\end{split}
\end{equation}
where $h_k\in\mathbb{R}^{n \times l}$ represents the hidden state at step $k$. $h_{k-1}\in\mathbb{R}^{n \times l}$ denotes the hidden state at step $k-1$, $A\in\mathbb{R}^{n\times n}$. $B\in\mathbb{R}^{n\times d}$, and $C\in\mathbb{R}^{n\times n}$ are the parameter matrices. $x_k\in\mathbb{R}^{d \times l}$ and $y_k\in\mathbb{R}^{n\times l}$ represent the input and output at step $k$, respectively. $l$, $d$, and $n$ denote the sequence length, input vector dimension, and state size, respectively

Recently, several significant developments have been made in SSM, aimed at enhancing their application in neural networks. G. Albert et al. introduced the time-step memory mechanism and HiPPO \cite{hippo}, leading to the development of the S4 model \cite{S4}. This addressed two major challenges in sequence modeling: the computational complexity associated with transformers \cite{attention_is,trandformers} and the sequence forgetting problem inherent in RNNs \cite{RNN}. Building upon insights from S5 \cite{smith2022simplified}, researchers proposed Mamba \cite{mamba}, which incorporates a hardware-aware algorithm and introduces a structured SSM gating mechanism \cite{SSM_in_control,ViM,mamba}. Subsequently, two major architectural improvements emerged: Vision Mamba \cite{ViM}, which focuses on structure blending, and VMamba \cite{vmamba}, which implements SS2D scanning. These innovations have established two primary directions for researchers working on vision-related tasks \cite{SSM_in_control,combine_cnn_with_ssm,mambamixer}.

In recent years, the Mamba architecture has emerged as a promising alternative to Transformers, offering reduced computational complexity through selective modeling and linear scaling. Mamba-based approaches have shown remarkable performance in various remote sensing tasks, including classification, semantic segmentation, change detection, and image super-resolution \cite{U-Mamba,Diffusion_Mamba,VideoMamba,zhang2024cdmamba}. Several methods have advanced this field: RSMamba \cite{chen2024rsmamba} introduced a multi-directional scanning mechanism; RSCaMa \cite{liu2024rscama} enhanced temporal change detection through cross-temporal scanning; ChangeMamba \cite{chen2024changemamba} combined global and local modeling approaches; RS3Mamba \cite{ma2024rs} explored local detail enhancement within the State Space Model (SSM); and Pan-Mamba \cite{he2024pan} advanced pan-sharpening technology while reducing redundancy. Despite these efforts, the research in instance segmentation remains limited. For example, MambaRSIS\cite{pan2025mambarsis} harnesses contextual information to facilitate multi-scale feature aggregation while investigating SSM blocks for efficient remote sensing image segmentation. The results are promising, demonstrating the potential of SSMs in remote sensing applications. Our work focuses on transferring segmentation priors from a promptable foundation model to a lightweight SSM student via knowledge distillation, and systematically studies RS-suitable design choices for efficient instance segmentation. To address this, we propose RS4D, a method specifically designed for efficient instance segmentation in remote sensing applications.

\section{Methodology}

This section introduces two key components of the proposed framework: first, the knowledge distillation mechanism that transfers pre-trained knowledge from a visual foundation model to a lightweight SSM, and second, the efficient instance segmentation method designed for remote sensing imagery.

\subsection{Linear-Time State Space Knowledge Distillation}

Our proposed method employs a knowledge distillation framework comprising a teacher model and a student model, with SSM as the core architecture. For the teacher model, we adopt the SAM Visual Encoder (ViT-Base), which has demonstrated superior feature extraction capabilities when trained on large-scale datasets \cite{SAM}. The student model encompasses three novel architectures specifically designed to optimize inference speed, enhance learning robustness, and minimize storage requirements while effectively distilling knowledge from the SAM-Encoder. These lightweight SSM-based architectures achieve an optimal balance between computational efficiency and model performance.

\subsubsection{SSM Student Architectures}

Our framework implements three distinct lightweight SSM student architectures: VanillaMamba, TransMamba, and ScanningMamba, as illustrated in Fig. \ref{fig:overall} (c). 

VanillaMamba consists of 12 stacked Vision Mamba blocks (ViM blocks). While this architecture achieves initial convergence during the first training stage, its simplified structure presents certain limitations. Notably, maintaining the original memory parameters while expanding the input resolution from $256 \times 256$ pixels to $1024 \times 1024$ pixels results in a $\times 16$ increase in sequence length, potentially constraining the model's learning capacity.

\begin{figure*}[t]
\centering
\includegraphics[width=1\linewidth]{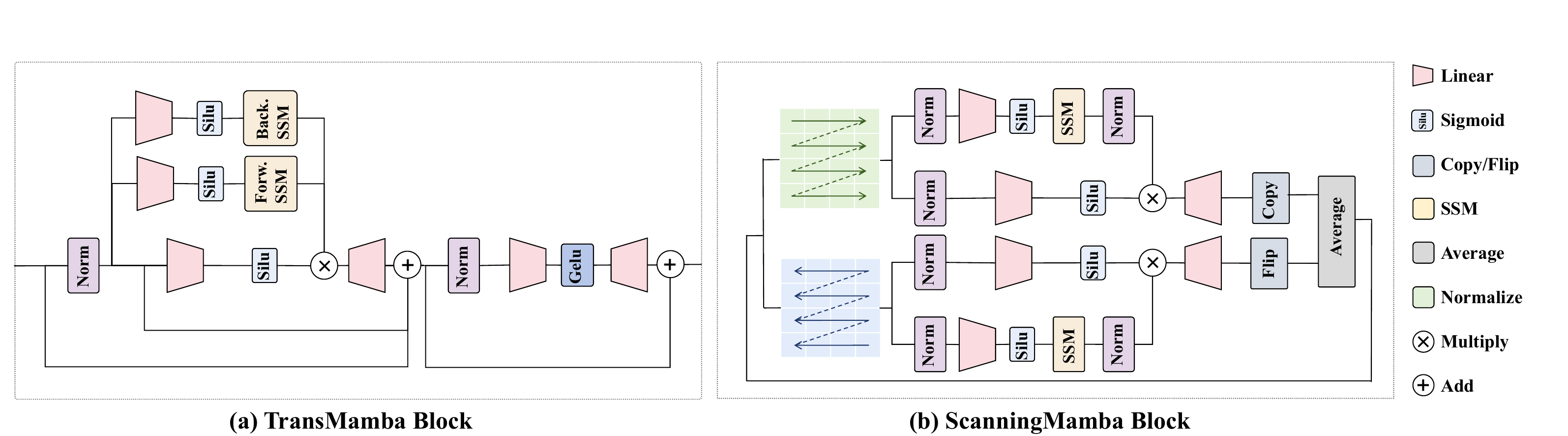}
\caption{
The architectural details of (a) TransMamba and (b) ScanningMamba blocks.
}
\label{fig:backbone2}
\end{figure*}

The designed TransMamba employs a hybrid Mamba-Transformer architecture, as depicted in Fig. \ref{fig:backbone2} (a). The architecture consists of 12 layers, each containing a specialized TRVim block that integrates Mamba components with the Transformer layer. The TRVim block implements a sequential combination of a Residual Layer, Normalization Layer, Vim Layer, GELU activation function, and a Multi-Layer Perceptron (MLP). The integration with the Transformer architecture is achieved through a selective replacement of the attention layer, rather than substituting the entire stack. Although TransMamba exhibits enhanced learning capabilities compared to VanillaMamba, its increased parameter count leads to higher computational complexity. Consequently, it is the most computationally intensive among the three architectures, requiring longer inference times. Experimental results indicate that while the model successfully converges for bounding box (BBox) detection, its segmentation performance remains suboptimal. The computational process can be formally expressed as:
\begin{equation}
\begin{split}
Y_{a-1}&=X_{a-1}+\text{TRVim}(\text{norm}(X_{a-1}))\\
X_{a}&=Y_{a-1}+\text{linear}_{mlp}(\text{Gelu}(\text{norm}(Y_{a-1})))
\end{split}
\end{equation}
where $X_{a} \in\mathbb{R}^{b\times l \times d}$ denotes the intermediate state at layer $a$, $\text{TRVim}(\cdot)$ represents the Mamba block within TransMamba, $a \in [1,12]$ represents the layer index, $\text{linear}_{mlp}(\cdot)$ signifies the MLP transformation, and $\text{Gelu}(\cdot)$ represents the GELU activation function.

ScanningMamba modifies the original Mamba architecture \cite{mamba} by implementing both forward and backward scanning strategies. When processing an input sequence, the ScanningMamba block generates two versions: one in the original order and another in reverse order. These sequences are processed independently, and then the reversed sequence is restored to its original order before being averaged with the forward-processed sequence. To better capture sequence complexity, we increased the number of stacking layers from 12 to 24. The detailed architecture is illustrated in Fig. \ref{fig:backbone2} (b). Despite its moderate size, ScanningMamba demonstrates robust learning capabilities and efficient inference. The process can be mathematically expressed as:
\begin{equation}
\begin{split}
X_{c}^{\text{backward}}&=\text{Mambamixer}(\text{flip}(X_{c-1}))\\
X_{c}^{\text{forward}}&=\text{Mambamixer}(X_{c-1})\\
X_{c}&=X_{c-1}+(X_{c}^{\text{backward}}+\text{flip}(X_{c}^{\text{forward}}))/2
\end{split}
\end{equation}
where $X_{c}\in\mathbb{R}^{b\times l \times d}$ represents the output of layer $c$. $X_{c-1}\in\mathbb{R}^{b\times l \times d}$ denotes the output of layer $c-1$. ${X}_{c}^{\text{backward}}\in\mathbb{R}^{b\times l \times d}$ and ${X}_{c}^{\text{forward}}\in\mathbb{R}^{b\times l \times d}$ are intermediate variables in layer $c$. $\text{Mambamixer}(\cdot)$ represents the Mambamixer function. $\text{flip}(\cdot)$ denotes the sequence reversal function. $c$ represents the layer number, ranging from 1 to 24.

Our three student backbone design is guided by a core question: how to achieve linear-time computation suitable for high-resolution remote sensing instance segmentation while retaining global segmentation priors similar to SAM. Since SAM itself is a 12-layer transformer, the VanillaMamba merely substitutes the transformer with the capabilities of the SSM block.

In contrast, TransMamba explores whether combining the structures of SSM and transformers can complement the long-range modeling of SSM. One natural idea is to integrate SSM with self-attention mechanisms, leveraging the global associations of self-attention alongside the selective modeling of SSM. However, retaining self-attention would not alter the computational complexity of the model, which contradicts our lightweight design principles. Therefore, we removed the self-attention layers from the transformer framework and replaced them with SSM blocks, keeping the skip connections to prevent excessive information compression from SSM, which could lead to the loss of shallow-layer information. Additionally, we replaced ReLU with GELU. Although ReLU was the original choice for the Transformer MLP, TransMamba is specifically designed for efficient and robust distillation in high-resolution remote sensing settings. This is particularly suitable for distillation in dense prediction scenarios. ReLU has truncation at zero, which can lead to gradient clipping and activation sparsity, causing all negative responses to be clipped to zero. In contrast, GELU is a smooth gated activation function, resulting in a more continuous backpropagation process, which reduces training-induced jitter. Additionally, many objects in remote sensing occupy only a tiny number of pixels in the overall high-resolution image, often accompanied by low-contrast complex backgrounds. Retaining some weak but useful texture or boundary cues can facilitate the refinement of small targets and boundary details. Therefore, we employ GELU in the lightweight MLP branch to enhance optimization robustness under masked and noise distillation inputs. It is important to note that in our mixed block, while the SSM path already provides a smooth gated nonlinear function (SiLU) intended to control information flow, we still need to offer a simple nonlinear supplement to encourage more stable activation of weak but important local responses, which is crucial in remote sensing images.

ScanningMamba tests whether there is a clear improvement in two-dimensional contextual completeness, specifically through feature aggregation via forward and backward scanning, which is a key factor for stable optimization and high mask quality. This design is motivated by the need for spatially consistent context aggregation in dense prediction tasks across the two-dimensional image plane. Pure one-dimensional ordering or unidirectional aggregation may introduce directional bias and incomplete context, which can adversely affect the consistency of masks in large high-resolution scenes. The memory mechanism of SSM is fundamentally based on its HIPPO matrix continuously compressing information. However, a single-path SSM inevitably biases the model towards remembering the most recently encountered information. Therefore, introducing multiple paths can effectively mitigate this limitation. ScanningMamba addresses this issue by employing bidirectional scanning on the two-dimensional token grid, achieving a more complete integration of spatial context while maintaining linear time complexity, thus enhancing mask quality and optimization stability. On the other hand, introducing too many paths can substantially increase the computational burden on the model. For this reason, we designed ScanningMamba and expanded it to 24 layers based on our experimental results, as too few layers in SSM are insufficient to carry the prior information of SAM. We have validated this in subsequent experiments.
\subsubsection{Distillation Process}

During knowledge distillation, the student network is trained using unlabeled data through the following process: First, 50,000 augmented images (10k raw images) are fed into both the teacher and student models. We deliberately introduce random masks and noise to the student model's input pixels to improve the generalization of features via self-supervised learning. The noise amplitude is adaptively adjusted based on the maximum value of the output matrix to maintain stability:
\begin{equation}
\begin{split}
X^{'}&=\text{mask}(X_{input})+{\alpha}_e \times \epsilon\\
{\alpha}_e&=\text{min}(\max_{b,l,d}(|Y^{e-1}|),\lambda_{const})\\
\epsilon &\sim \text{Uniform}(0, 1)\\
\end{split}
\end{equation}
where $X^{'}\in\mathbb{R}^{b\times l \times d}$ denotes the student's inputs at the current epoch, and $X_{input}\in\mathbb{R}^{b\times l \times d}$ represents the original image. The function $\text{mask}(\cdot)$ applies a binary mask to the input. $Y^{e-1} \in \mathbb{R}^{b\times l \times d}$ represents the student's output from the previous epoch, and $\max_{b,l,d}(|Y^{e-1}|)$ is a scalar, the maximum absolute value over all elements of that output tensor. ${\lambda}_{const}$ is a constant threshold. The functions $\text{min}()$ and $\text{max}()$ return the minimum and maximum elements, respectively. The term $\alpha$ represents the adaptive noise amplitude in the current epoch, while $\epsilon$ is sampled from a uniform distribution. In implementation, we do not store the full student outputs from previous epochs on disk. Instead, $\max_{b,l,d}(|Y^{e-1}|)$ is computed online as a running maximum over mini-batches during epoch ${e-1}$, and only this scalar is retained for epoch $e$. Therefore, the additional memory overhead is negligible.

The supervision signal is derived from the consistency between the teacher and student networks' outputs, measured by their L1 distance, which is formulated as:
\begin{equation}
\begin{split}
    \mathcal{L}_{kd}=\frac{1}{b\times l \times d} \sum_{i=1}^{l} \sum_{j=1}^{b} | {y_{ij}}-{y_{ij}^{\text{pred}}}|
\end{split}
\end{equation}
where $y_{ij}\in\mathbb{R}^{d}$ and $y_{ij}^{\text{pred}}\in\mathbb{R}^{d}$ denote the outputs of the teacher and student networks, respectively, for the $i$-th token in the $j$-th batch. The indices $i$ and $j$ are integers ranging from 1 to $l$ and 1 to $b$, respectively.

It is important to note that although SA-1B\cite{SAM} is a natural-image dataset, our goal is to distill SAM into a student model that transfers well to remote sensing imagery, where the data distribution differs substantially. Remote sensing images typically have higher spatial resolution, different sensor characteristics, and more complex high-frequency textures. To reduce this domain gap during distillation, we introduce an RS-motivated augmentation that injects high-frequency perturbations into SA-1B\cite{SAM} images. Specifically, we add a mixture of noise and masks to emphasize high-frequency components with randomly sampled intensities. This approach encourages the student to learn robust, shape- and structure-oriented representations rather than overfitting to low-frequency appearance cues in natural images, thereby improving its adaptation to remote sensing imagery. Relevant experiments validate the reasonableness of this approach. We apply this perturbation only during distillation (pretraining), while keeping the downstream RS fine-tuning unchanged.

\begin{figure*}
	\centering
	\includegraphics[width=1\linewidth]{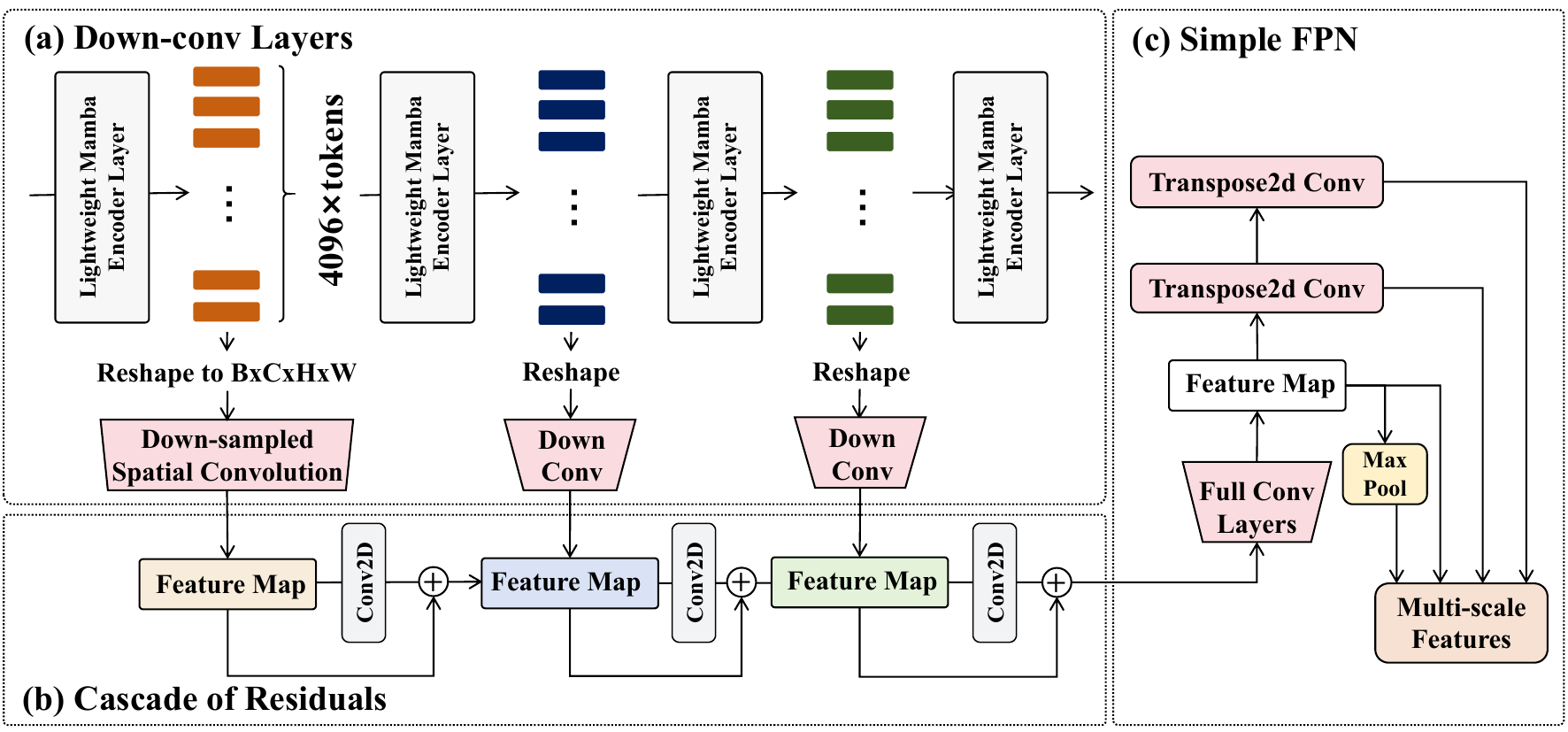}
	\caption{{The structure of neck part of the proposed model. The neck consists of three parts: (a) a multi-level token feature downsampling module, (b) a cascaded residual module, and (c) a simple FPN module.} 
 }
	\label{fig:neck}
\end{figure*}

\subsection{Efficient Remote Sensing Instance Segmentation}

The efficient segmentation model consists of three primary components: the backbone, the neck, and the head. The backbone architecture is derived from previously discussed knowledge distillation techniques. The neck component performs multi-level feature fusion, while the head component executes task-specific operations.

\subsubsection{Multi-level Feature Fusion Neck}

The neck serves as the connecting component between the backbone and head, extracting high-level semantic information through feature abstraction. As illustrated in Fig. \ref{fig:neck}, it comprises three key components: a downsampled feature layer, cascade of residuals, and SimpleFPN. The downsampled feature layer reshapes sequences from odd layers while performing spatial downsampling. To preserve shallow features for edge detail refinement, the cascade of residuals network implements skip connections between deep and shallow feature layers. The SimpleFPN generates standard multi-scale feature maps through cascading downsampling and upsampling convolutions. 

We fuse multi-depth features and build multi-scale maps at the final embedding stage. Unlike CNN backbones, SSM/ViT backbones typically operate on a single token/grid resolution and do not naturally produce an explicit multi-resolution feature pyramid across stages.  Constructing a progressive FPN pyramid by injecting features into multiple intermediate upsampling layers would require additional down/up-sampling operators and repeated fusion at several levels, increasing computation and undermining our goal of a lightweight, linear-time design.  Instead, we aggregate complementary information from multiple depths at a unified resolution via lightweight skip connections, and then use a minimal SimpleFPN to output standard multi-scale feature maps for the detector. This multi-level feature fusion neck operates on minimal spatial and channel dimensions, resulting in highly efficient computational costs.

Remote sensing images often contain sparse small objects, ranging from small vehicles to ships, and do not exhibit the abundance of large-scale instances found in natural images, such as those in the COCO dataset, where one object may cover a significant portion of the image. Since our backbone design incorporates dense patches, this allows the model to focus more on smaller objects. Therefore, when designing a neck that complements this backbone, we prefer to strategically select and fuse features from specific intermediate layers of the SSM backbone, rather than structuring the model like a ResNet + FPN to extract and merge multi-scale features. This approach enhances the model's feature extraction capability in remote sensing images compared to natural images.

\subsubsection{Instance Segmentation Head}

\begin{figure*}
	\centering
	\includegraphics[width=1\linewidth]{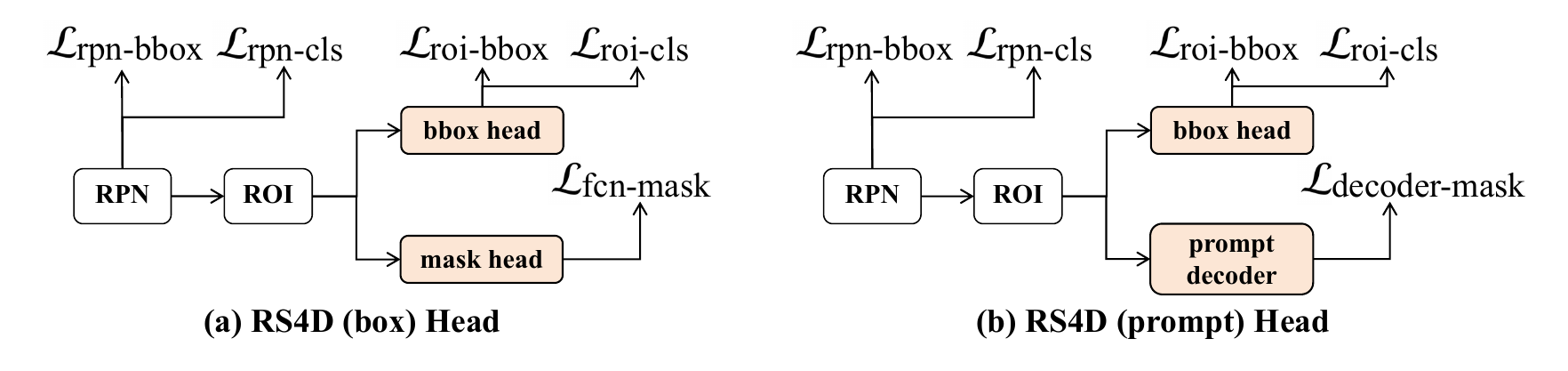}
	\caption{{The structure of the head part of the proposed model. We show two different ways of implementation of the network's head part for instance segmentation: (a) box head and (b) prompt head. } 
 }
	\label{fig:head}
\end{figure*}

As illustrated in Fig. \ref{fig:head}, the segmentation head is designed with two variants: a box-based head and a prompt-based head. The box-based head follows the approach of Mask R-CNN \cite{Mask_r-cnn}, incorporating a Region Proposal Network (RPN) head and a Region of Interest (RoI) head. The total loss function is formulated as:
\begin{equation}
    \begin{split}
        \mathcal{L} = \mathcal{L}_{rpn-cls}&+ \mathcal{L}_{rpn-bbox}+ \mathcal{L}_{roi-cls}\\
        &+\mathcal{L}_{roi-bbox}+ \mathcal{L}_{fcn-mask}
    \end{split}
\end{equation}
where $\mathcal{L}_{rpn-cls}$ and $\mathcal{L}_{rpn-bbox}$ denote the classification and bounding box losses of the RPN, respectively; $\mathcal{L}_{roi-cls}$ and $\mathcal{L}_{roi-bbox}$ represent the Region of Interest (RoI) classification and regression losses, respectively; and $\mathcal{L}_{fcn-mask}$ represents the binary cross-entropy mask loss computed by a simple Fully Convolutional Network (FCN).

The prompt-based head eliminates the FCN component present in the box-based head and instead feeds the RoI features directly into the SAM prompt encoder and decoder to generate binary masks. The corresponding loss function is defined as:
\begin{equation}
    \begin{split}
        \mathcal{L} = \mathcal{L}_{rpn-cls}&+ \mathcal{L}_{rpn-bbox}+ \mathcal{L}_{roi-cls}\\
        &+\mathcal{L}_{roi-bbox}+ \mathcal{L}_{decoder-mask}
    \end{split}
\end{equation}
where $\mathcal{L}_{decoder-mask}$ represents the loss between the decoder-generated segmentation output and the ground-truth mask.

It should be noted that the Box Head adheres to a traditional Mask R-CNN design, where RoI-aligned features are input into a fully convolutional network (FCN) mask predictor to produce the instance mask. In contrast, the Prompt Head replaces the FCN mask predictor with a procedure that employs SAM-style prompt encoding and mask decoding. In this approach, RoI-aligned features are first mapped into a small set of learned prompt tokens, which act as implicit prompts derived from the detected RoI region rather than from user interactions like clicks or bounding boxes. Thus, in our context, prompt specifically refers to the learned prompt tokens constructed from RoI features, rather than manually provided prompts.

\subsection{Implementation Details}

\subsubsection{Network Design}

RS4D is implemented using MMDetection \cite{chen2019mmdetection} and MMEngine \cite{mmengine2022} frameworks, which organize the model architecture into three primary components: the backbone, the neck, and the head. Among the tested backbone variants, we focus on ScanningMamba due to its superior convergence rate and performance in our preliminary experiments. The neck architecture adopts the feature aggregator design from RSPrompter \cite{RSPrompter}, with modifications only to the number of aggregator layers.

\subsubsection{Training Details}
\label{TD}

Following the model's structural design, the training process consists of two main phases: knowledge distillation of the backbone and fine-tuning of the complete model for instance segmentation tasks.

For backbone distillation, unless otherwise specified, we utilize 0.1\% of the unlabeled data from the readily available SA-1B dataset\cite{SAM}, which consists of 10k images (0.1\% of SA-1B\cite{SAM}). For the NWPU experiments, we use distilled weights obtained from 10\% (1000k) of SA-1B to mitigate the data scarcity of downstream training. In distillation, we choose to use only a subset for distillation for two main reasons. First, it allows us to explore computational and storage efficiency while keeping the overall transfer process low-cost. Importantly, even this small subset is sufficient to convey meaningful segmentation priors from the SAM encoder to the lightweight SSM student, aligning with our goal of efficient knowledge transfer. The second reason is that the complete SA-1B dataset\cite{SAM} contains over 11 million images, which incurs substantial training and storage costs, comparable to pre-training models using MAE, and contradicts our original intention. Note that 10k images refers to the number of unique SA-1B images in the subset. The SA-1B subsets are constructed directly following the official data organization and download order. During distillation, we apply stochastic augmentations multiple times, resulting in roughly 50k augmented training samples. We emphasize that the 0.1\% setting should be viewed as an efficient default rather than a universal saturation point, and the optimal distillation data volume depends on the downstream dataset regime.

\begin{figure}
\centering
\includegraphics[width=1\linewidth]{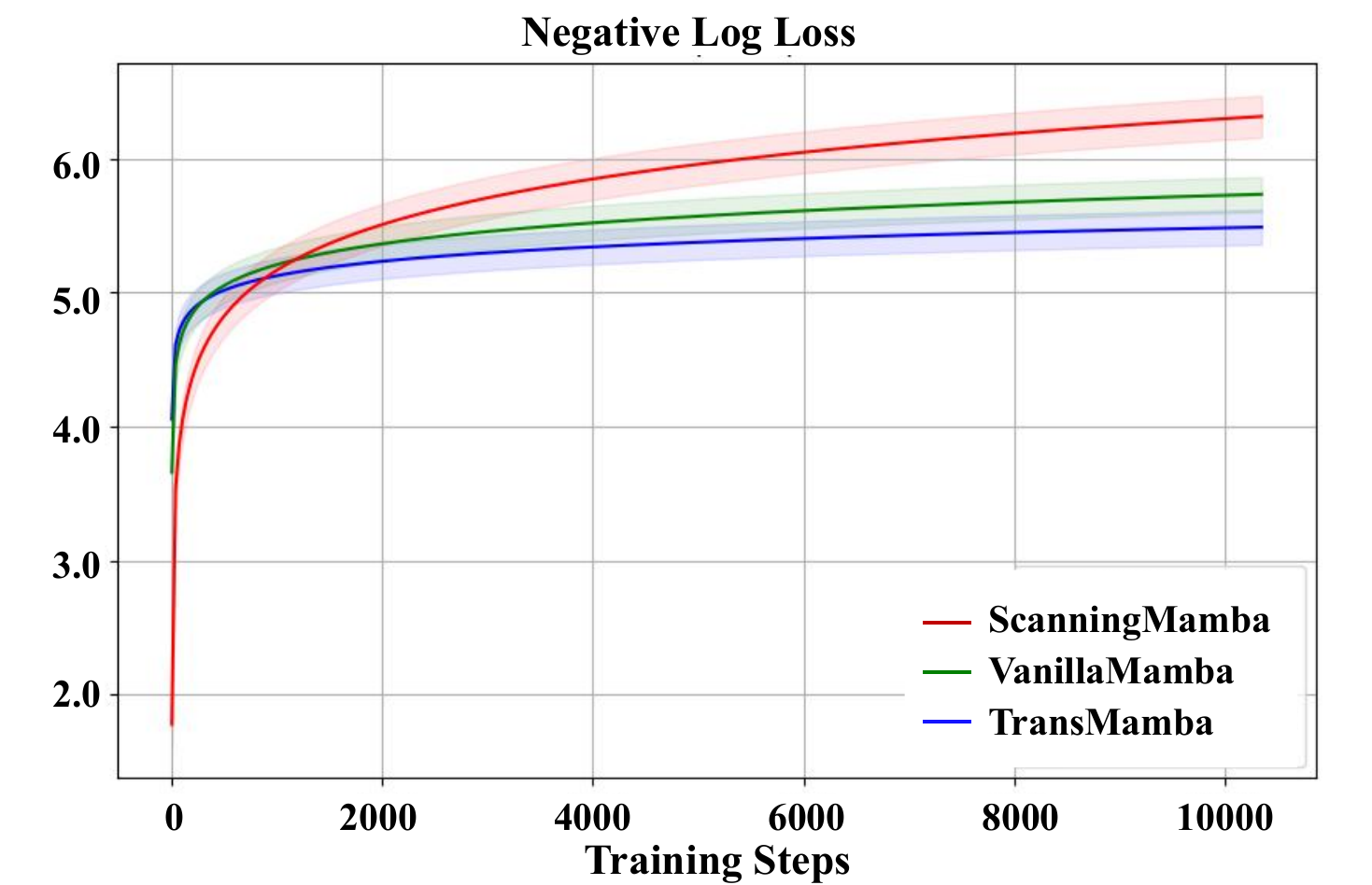}
\vspace{-4ex}
\caption{
Comparison of backbone architectures in terms of the negative log loss and training steps, where higher values indicate faster convergence. The shaded region around each curve represents the loss variance calculated using a sliding window, providing insight into training stability.
}
\label{fig:backbone3}
\end{figure}

Although there is a domain gap between this dataset and remote sensing imagery, this discrepancy has a limited effect on downstream tasks, as we are mainly transferring general knowledge. All images are resized to 1024$\times$1024 resolution and fed into the teacher model. Image flipping is also used as a data augmentation technique. Simultaneously, masked and noise-augmented versions of the same images are input to the SSM student model for feature extraction. L1 distance is applied to the final outputs to ensure that the deep features extracted by both models remain closely aligned. As demonstrated in Fig. \ref{fig:backbone3}, ScanningMamba achieves the optimal balance between performance and efficiency, along with more stable convergence. This indicates that bidirectional scanning enhances the completeness of features for dense mask prediction while maintaining the linear-time advantage of SSM. Consequently, we have selected ScanningMamba as the default backbone for the subsequent fine-tuning phase.

The instance segmentation model's backbone is initialized using the distilled SSM, and all model parameters are fine-tuned on the specific remote sensing instance segmentation dataset. The model is trained for 800 epochs. We implement basic data augmentation techniques, including random flipping, resizing, and cropping. For optimization, we employ the AdamW optimizer with an initial learning rate of $1\times10^{-4}$ and a weight decay of 0.05. The learning rate is adjusted using a cosine annealing scheduler with linear warm-up. Training is conducted using bfloat16 precision. Other parameters use the default configuration of MMDetection\cite{chen2019mmdetection}, considering the robustness of the framework, and for the sake of model reproducibility, while reducing training costs. We did not use multiple random seeds for averaging. In the distillation phase, we built the code using native PyTorch, with a random seed set to 42.

\section{Experiments and Analysis}

\subsection{Experimental Setup}

\subsubsection{Dataset and Evaluation Metrics}

We fine-tune and evaluate RS4D on three publicly available remote sensing instance segmentation datasets: the SAR Ship Detection Dataset (SSDD) \cite{SSDD}, the Building Extraction Dataset (WHU) \cite{ji2018fully}, and the NWPU VHR-10 Dataset (NWPU) \cite{NWPU}. All experimental settings are consistent with those used in RSPrompter.

SSDD \cite{SSDD} is a public SAR ship dataset that primarily contains small targets. The dataset consists of $500 \times 500$ pixel images with spatial resolutions ranging from 1 to 15 meters. The targets are annotated using horizontal bounding boxes.

WHU \cite{ji2018fully} is a comprehensive aerial building dataset containing high-quality imagery. The dataset comprises $512 \times 512$ pixel images with a spatial resolution of 0.3 meters. It is divided into 4,736 training images, 1,036 validation images, and 2,416 testing images.

NWPU \cite{NWPU} encompasses ten object categories: ships, aircraft, ports, bridges, oil tanks, baseball fields, tennis courts, basketball courts, track and field facilities, and cars. The dataset is relatively small, consisting of 650 target images and 150 background images.

Due to the relatively small size of the NWPU dataset, along with its significant variations in target scale and the number of categories, we initialize the backbone using distilled weights obtained from a larger subset of SA-1B (10\%) and employed test time augmentation during testing.
\begin{figure*}
\centering
\includegraphics[width=1\linewidth]{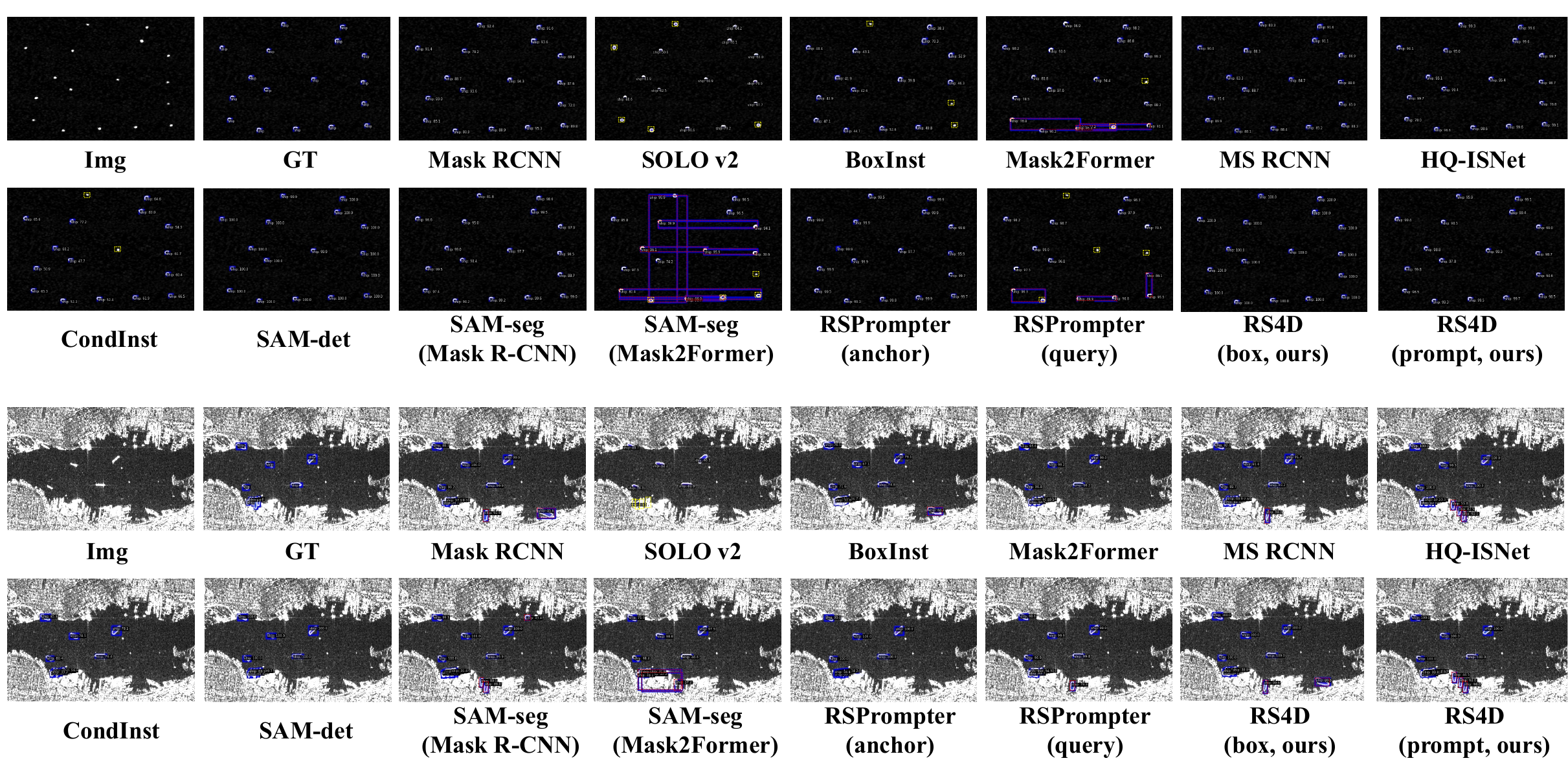}
\caption{{Images between the proposed method and some SOTA instance segmentation methods on the infrared ship instance segmentation dataset SSDD \cite{SSDD}. Red bounding boxes represent false positives, while yellow bounding boxes indicate false negatives.} 
}
\label{fig:ssdd1}
\end{figure*}

To comprehensively evaluate the performance of the proposed method, we adopt two sets of metrics to assess the model's accuracy and efficiency. For accuracy, we use mean Average Precision (mAP), specifically $\text{AP}_{\text{bbox}}$ for bounding boxes, $\text{AP}_{\text{mask}}$ for instance masks, and $\text{AP}^\text{50}$ and $\text{AP}^\text{75}$ for IoU thresholds of 0.5 and 0.75, respectively. For efficiency, we measure the Floating-point Operations (FLOPs) and the number of parameters.

\subsection{Qualitative Analysis}

We compare the proposed RS4D with other state-of-the-art approaches, including traditional instance segmentation models such as Mask R-CNN \cite{Mask_r-cnn}, Mask Scoring R-CNN \cite{Mask_scoring_r-cnn}, HTC \cite{HTC}, SOLOv2 \cite{solov2}, CondInst \cite{Coninst}, BoxInst \cite{boxinst}, Mask2Former \cite{maskformer}, and Cascade Mask RCNN \cite{cai2019cascade}. We also compared the typical models for natural images, including SwinV2 (HTC++) \cite{liu2022swin}, CBNetv2 \cite{liang2022cbnet}, and ViTDet \cite{li2022exploring}. Additionally, we compare against specialized remote sensing models including CATNet \cite{CATNet} and HQ-ISNet \cite{HQ-ISNet}, as well as foundation model variants \cite{RSPrompter}: SAM-cls, SAM-det, SAM-seg, RSPrompter (anchor), and RSPrompter (query). It is important to note that although  SwinV2 (HTC++) \cite{liu2022swin} and ViTDet \cite{li2022exploring} are not pretrained foundation models like SAM, they also use ViT as their backbone, and have a comparable or similar number of parameters to the foundational model.

During training, the model's training configuration remains consistent, as detailed in Subsection III.C \nameref{TD}. In the ablation experiments, all configurations except for the ablated variables are kept the same. The variables that are changed are the masking ratio and the noise ratio used in pretraining, both of which are selected from 0\%, 10\%, 20\%, and 30\%.

\subsubsection{Comparative Analysis on the SSDD Dataset}

\begin{table*}[!t]  
\caption{Comparison with other methods on the SSDD dataset \cite{SSDD}.}
\label{tab:Comparisons_other_methods_ssdd}
\centering
\resizebox{\linewidth}{!}{
\begin{tabular}{c|c|c c c c c c c c c}
	\toprule
 \multicolumn{1}{c|}{\shortstack{Method}} & $\text{Backbone}$&$\text{AP}_{\text{bbox}}$ & $\text{AP}^{50}_{\text{bbox}}$ & $\text{AP}^{75}_{\text{bbox}}$ & $\text{AP}_{\text{mask}}$ & $\text{AP}^{50}_{\text{mask}}$ & $\text{AP}^{75}_{\text{mask}}$&$\text{Params (M)}$&$\text{bb-Params (M)}$&$\text{bb-FLOPs (G)}$\\
        \midrule
 \multicolumn{1}{c|}{Mask RCNN \cite{Mask_r-cnn}}&{ResNet50} & 67.7 & 95.6 & 84.9 & 64.3 & 92.6 & 80.9 & 44.18&\cellcolor{gray}25.56&172.69\\

 \multicolumn{1}{c|}{MS RCNN \cite{Mask_scoring_r-cnn}}&{ResNet50} & 67.8 & 95.3 & 85.9 & 64.9 & 93.3 & 80.4 &60.74&\cellcolor{gray}25.56&172.69\\

 \multicolumn{1}{c|}{HTC \cite{HTC}}&{ResNet50} & 69.3 & 97.1 & 85.7 & 64.1 & 94.4 & 80.6 &46.59&\cellcolor{gray}25.56&172.69\\

 \multicolumn{1}{c|}{SOLO v2 \cite{solov2}}&{ResNet50} & - & - & - & 58.5 & 86.2 & 74.0&46.58&\cellcolor{gray}25.56&172.69\\

 \multicolumn{1}{c|}{SCNet \cite{Scnet}}&{ResNet50} & 66.9 & 92.5 & 82.5 & 64.9 & 92.6 & 80.1&94.80&\cellcolor{gray}25.56&172.69\\

 \multicolumn{1}{c|}{CondInst \cite{Coninst}}&{ResNet50} & 68.1 & 92.4 & 85.5 & 62.5 & 93.4 & 81.2&34.16&\cellcolor{gray}25.56&172.69\\

 \multicolumn{1}{c|}{BoxInst \cite{boxinst}}&{ResNet50} & 62.8 & 96.2 & 74.7 & 45.2 & 92.3 & 35.3&35.14&\cellcolor{gray}25.56&172.69\\

 \multicolumn{1}{c|}{Mask2Former \cite{maskformer}}&{ResNet50} & {62.7} & 90.7 & 75.6 & 64.4 & 93.0 & 82.4 &44.02&\cellcolor{gray}25.56&172.69\\

 \multicolumn{1}{c|}{CATNet \cite{CATNet}}&{ResNet50} & {67.5} & 96.8 & 80.4 & 63.9 & 93.7 & 80.1 &54.77&\cellcolor{gray}25.56&172.69\\

 \multicolumn{1}{c|}{HQ-ISNet \cite{HQ-ISNet}}&{ResNet50} & {66.6} & 95.9 & 80.2 & 63.4 & 95.7 & 78.1&80.56&\cellcolor{gray}25.56&172.69\\
 \multicolumn{1}{c|}{Cascade Mask RCNN \cite{cai2019cascade}}&{ResNet50} & \cellcolor{gray}{70.4} &\cellcolor{gray}97.5 & 85.9 & \cellcolor{gray}67.9 & \cellcolor{gray}96.6 & \cellcolor{gray}85.5&77.02&\cellcolor{gray}25.56&172.69\\
 \midrule
\multicolumn{1}{c|}{SwinV2 (HTC++) \cite{liu2022swin}}&{SwinV2-Huge} & {59.4} & 85.5 &  76.8 & {56.4} & {85.6} & {70.0}&709.09&655.14&303.37\\
   \multicolumn{1}{c|}{CBNetv2 \cite{liang2022cbnet}}&{Swin-Tiny} & {40.6} & 62.7 &  48.5 & {38.6} & {62.5} & {46.8}&75.58&28.30&\cellcolor{darkgray}12.47\\
  \multicolumn{1}{c|}{ViTDet \cite{li2022exploring}}&{Swin-Huge} & {62.4} & 95.1 &  73.0 & {61.4} & {94.9} & {77.2}&679.42&655.58&375.10\\
 \multicolumn{1}{c|}{SAM-seg (Mask RCNN) \cite{RSPrompter}}&{ViT-Base} & {68.7} & 97.2 & 84.3 & 66.1 & 94.5 & 83.7&309.69&89.67&972\\

 \multicolumn{1}{c|}{SAM-seg (Mask2Former) \cite{RSPrompter}}&{ViT-Base} & {63.0} & 94.9 & 75.6 & 66.5 & 95.0 & 83.6&112.16&89.67&972 \\

 \multicolumn{1}{c|}{SAM-cls \cite{RSPrompter}}&{ViT-Base} & {43.2} & 70.8 & 48.8 & 47.5 & 78.1 & 57.7&{-}&{-}&{-} \\

 \multicolumn{1}{c|}{SAM-det \cite{RSPrompter}}&{ViT-Base} & {70.0} & 95.8 & 85.3 & 46.0 & 93.8 & 37.0&135.08&89.67&972\\

 \multicolumn{1}{c|}{RSPrompter \cite{RSPrompter}(anchor)}&{ViT-Base} &\cellcolor{gray} {70.4} &\cellcolor{darkgray}97.7 & \cellcolor{gray}86.2 & 66.8 & 94.7 & 84.0 &117.09&89.67&972\\

 \multicolumn{1}{c|}{RSPrompter \cite{RSPrompter}(query)}&{ViT-Base} & {66.0} & 95.6 & 78.7 & 67.3 &95.6 & 84.3&101.02&89.67&972 \\
 \midrule

 \multicolumn{1}{c|}{RS4D (box) }&{SSM}&\cellcolor{darkgray}70.7&97.4&\cellcolor{darkgray}86.4&\cellcolor{darkgray}70.0&\cellcolor{darkgray}97.4&\cellcolor{darkgray}87.5&\cellcolor{darkgray}33.20&\cellcolor{darkgray}11.75&\cellcolor{gray}108.8\\

 \multicolumn{1}{c|}{RS4D (prompt)}&{SSM}&69.7&96.6&83.2&64.7&93.7&80.9&\cellcolor{gray}38.83&\cellcolor{darkgray}11.75&\cellcolor{gray}108.8\\

	\bottomrule
\end{tabular}
}
\end{table*}

The performance comparison between RS4D and other methods on the SSDD dataset is presented in Tab. \ref{tab:Comparisons_other_methods_ssdd}. The shaded cells indicate the top two performing models, with darker colors representing higher rankings. RS4D outperforms both traditional approaches and recent foundation models, achieving 70.7 $\text{AP}_{\text{bbox}}$ and 70.0 $\text{AP}_{\text{mask}}$ with its box-based variant. The prompt-based RS4D demonstrates slightly lower performance than its box-based counterpart, yet still ranks fifth in detection metrics among the 22 evaluated methods. Furthermore, while maintaining a similar parameter count to CondInst and BoxInst, RS4D demonstrates significantly superior performance. Despite the substantial modality difference between SAR data and the SA-1B dataset\cite{SAM}, experimental results indicate that our method exhibits robust generalization capabilities and high efficiency even when transferred to a limited dataset.

\subsubsection{Comparative Analysis on the WHU Dataset}

\begin{figure*}
\centering
\includegraphics[width=1\linewidth]{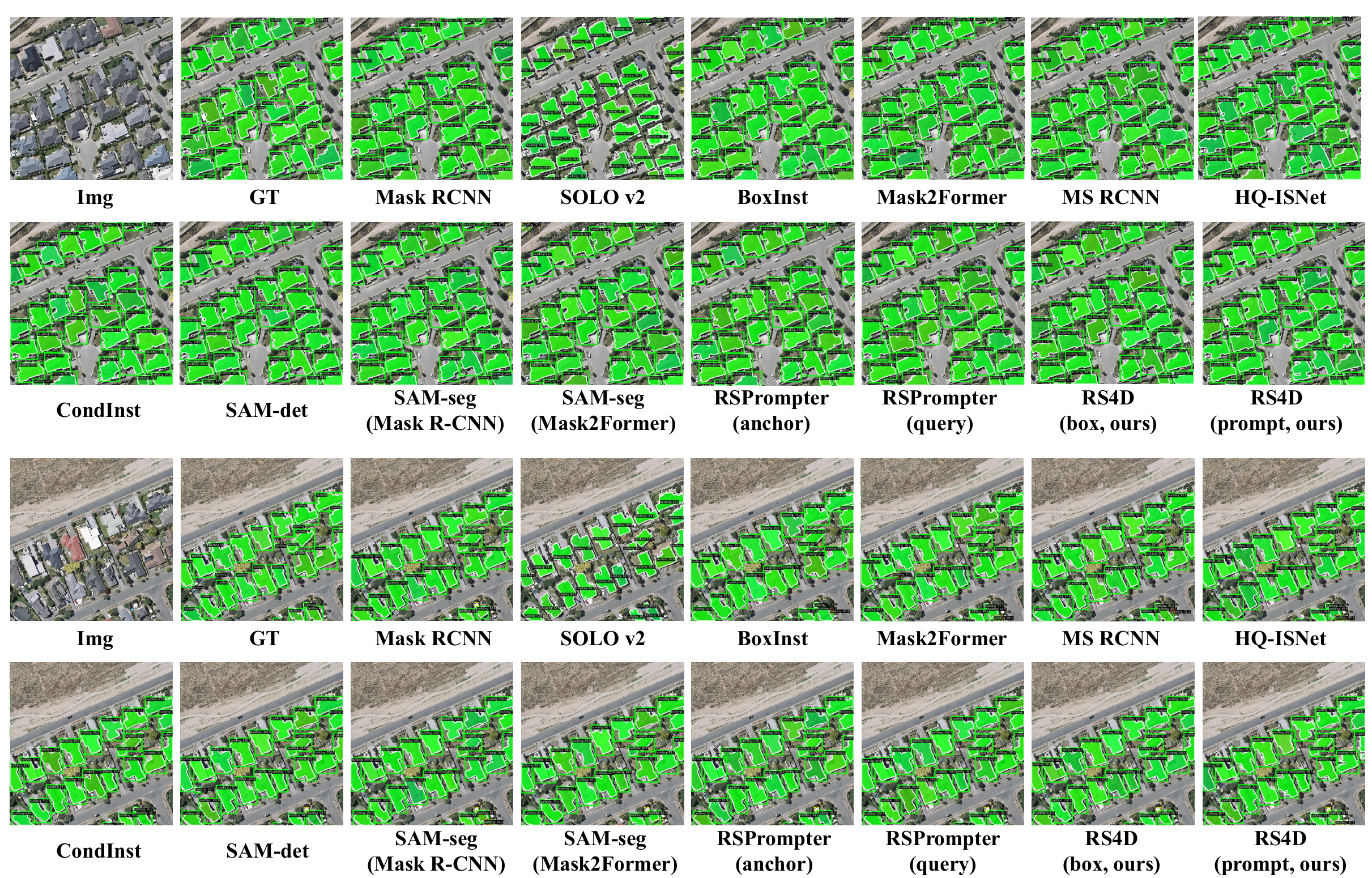}
\caption{{Images between the proposed method and some SOTA instance segmentation methods on the building instance segmentation dataset WHU \cite{ji2018fully}. Red bounding boxes represent false positives, while yellow bounding boxes indicate false negatives.} 
}
\label{fig:whu2}
\end{figure*}

\begin{figure*}
\centering
\includegraphics[width=1\linewidth]{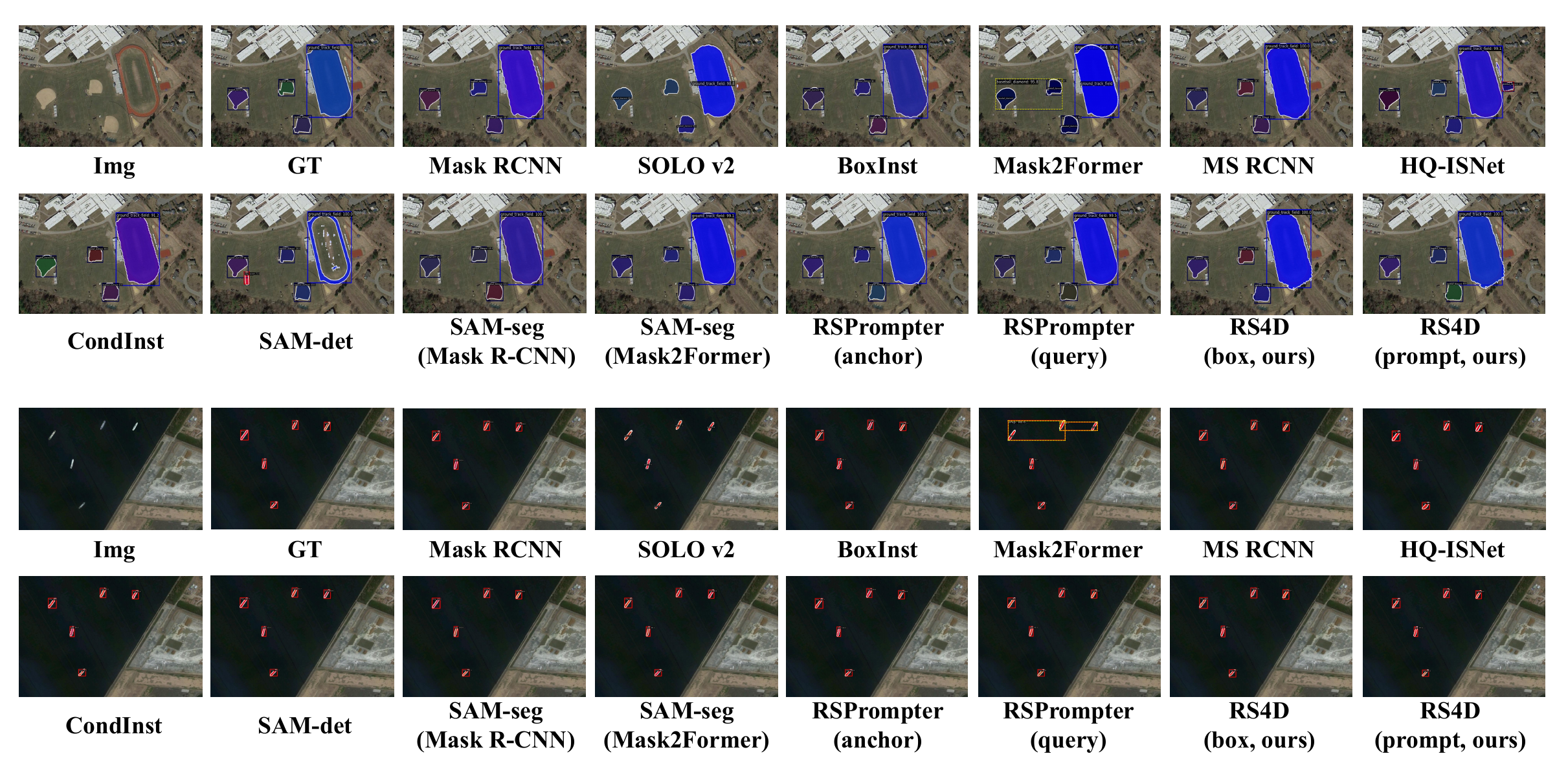}
\caption{{Images between the proposed method and some SOTA instance segmentation methods on the instance segmentation dataset NWPU \cite{NWPU}. Red bounding boxes represent false positives, while yellow bounding boxes indicate false negatives.} 
}
\label{fig:nwpu1}
\end{figure*}

\begin{table*}[!t]  
\caption{Comparison with other methods on the WHU dataset \cite{ji2018fully}.}
\label{tab:Comparisons_other_methods_whu}
\centering
\resizebox{\linewidth}{!}{
\begin{tabular}{c|c|c c c c c c c c c}
	\toprule
 \multicolumn{1}{c|}{\shortstack{Method}}& $\text{Backbone}$ & $\text{AP}_{\text{bbox}}$ & $\text{AP}^{50}_{\text{bbox}}$ & $\text{AP}^{75}_{\text{bbox}}$ & $\text{AP}_{\text{mask}}$ & $\text{AP}^{50}_{\text{mask}}$ & $\text{AP}^{75}_{\text{mask}}$&Params (M)&$\text{bb-Params (M)}$&$\text{bb-FLOPs (G)}$\\
        \midrule
 \multicolumn{1}{c|}{Mask RCNN \cite{Mask_r-cnn}}&{ResNet50} & 66.4 & 86.3 & 76.3 & 65.6 & 87.1 & 76.7&44.18&\cellcolor{gray}25.56&172.69 \\
 \multicolumn{1}{c|}{MS RCNN \cite{Mask_scoring_r-cnn}}&{ResNet50} & 67.7 & 87.2 & 77.1 & 66.9 & 87.5 & 77.5&60.74&\cellcolor{gray}25.56&172.69 \\
 \multicolumn{1}{c|}{HTC \cite{HTC}}&{ResNet50} & 68.4 & 87.8 & 77.8 & 67.7 & 88.1 & 78.3&46.59&\cellcolor{gray}25.56&172.69 \\
 \multicolumn{1}{c|}{SOLO v2 \cite{solov2}}&{ResNet50} & - & - & - & 65.2 & 86.7 & 75.2&46.58&\cellcolor{gray}25.56&172.69 \\
 \multicolumn{1}{c|}{SCNet \cite{Scnet}}&{ResNet50} & 68.1 & 87.6 & 77.7 & 66.5 & 87.9 & 81.2&94.80&\cellcolor{gray}25.56&172.69 \\
 \multicolumn{1}{c|}{CondInst \cite{Coninst}}&{ResNet50} & 66.7 & 87.5 & 76.7 & 66.6 & 87.8 & 78.7&34.16&\cellcolor{gray}25.56&172.69 \\
 \multicolumn{1}{c|}{BoxInst \cite{boxinst}}&{ResNet50} & 66.7 & 86.4 & 75.7 & 55.0 & 86.5 & 63.2&35.14&\cellcolor{gray}25.56&172.69  \\
 \multicolumn{1}{c|}{Mask2Former \cite{maskformer}}&{ResNet50} & {69.3} & 90.7 & 75.6 & 64.4 & \cellcolor{darkgray}93.0 & \cellcolor{gray}82.4&44.02&\cellcolor{gray}25.56&172.69 \\
 \multicolumn{1}{c|}{CATNet \cite{CATNet}}&{ResNet50} & {66.7} & 86.3 & 76.4 & 66.1 & 86.6 & 76.8&54.77&\cellcolor{gray}25.56&172.69 \\
 \multicolumn{1}{c|}{HQ-ISNet \cite{HQ-ISNet}}&{ResNet50} & {66.1} & 86.0 & 75.7 & 66.5 & 86.4 & 78.9&80.56&\cellcolor{gray}25.56&172.69 \\
  \multicolumn{1}{c|}{Cascade Mask RCNN \cite{cai2019cascade}}&{ResNet50} & {65.3} & 83.6 & 75.1 & 64.5 & 84.5 & 75.4 &77.02&\cellcolor{gray}25.56&172.69\\
 \midrule
\multicolumn{1}{c|}{SwinV2 (HTC++) \cite{liu2022swin}}&{SwinV2-Huge} & {46.8} & 62.2 &  54.5 & {46.7} & {62.5} & {56.2}&709.09&655.14&303.37 \\
\multicolumn{1}{c|}{CBNetv2 \cite{liang2022cbnet}}&{Swin-Tiny} & {60.0} & 86.6 &  73.6 & {58.8} & {85.8} & {74.6}&75.58&28.30&\cellcolor{darkgray}12.47 \\
\multicolumn{1}{c|}{ViTDet \cite{li2022exploring}}&{Swin-Huge} & {61.1} & 82.9 &  71.2 & {59.3} & {83.1} & {69.9}&679.42&655.58&375.10 \\
 \multicolumn{1}{c|}{SAM-seg (Mask RCNN) \cite{RSPrompter}}&{ViT-Base} & {70.3} & 89.8 & 81.9 & 70.1 & 89.9 & 81.0&309.69&89.67&972 \\
 \multicolumn{1}{c|}{SAM-seg (Mask2Former) \cite{RSPrompter}}&{ViT-Base} & {70.7} & 88.4 & 79.1 & \cellcolor{gray}71.1 & 89.5 & 81.1&112.16&89.67&972 \\
 \multicolumn{1}{c|}{SAM-cls \cite{RSPrompter}}&{ViT-Base} & {46.8} & 65.7 & 53.5 & 49.3 & 71.2 & 57.6&{-}&{-} \\
 \multicolumn{1}{c|}{SAM-det \cite{RSPrompter}}&{ViT-Base} & {69.1} & \cellcolor{gray}90.1 & 79.2 & 61.8 & 89.1 & 71.0&135.08&89.67&972\\
 \multicolumn{1}{c|}{RSPrompter (anchor) \cite{RSPrompter}}&{ViT-Base} & \cellcolor{gray}{71.9} & 90.0 & \cellcolor{darkgray}82.8 & 70.4 & 90.0 & 80.5&117.09&89.67&972 \\
 \multicolumn{1}{c|}{RSPrompter (query) \cite{RSPrompter}}&{ViT-Base} & \cellcolor{darkgray}{72.5} & \cellcolor{darkgray}91.0 & \cellcolor{gray}81.7 & \cellcolor{darkgray}72.5 & \cellcolor{gray}92.0 & \cellcolor{darkgray}82.9&101.02&89.67&972 \\
   \midrule
 \multicolumn{1}{c|}{RS4D (box)}&{SSM} & {70.3} & 89.3 & 79.2 & 70.1 & 89.5 & 80.3&\cellcolor{darkgray}33.20&\cellcolor{darkgray}11.75&\cellcolor{gray}108.8\\
 \multicolumn{1}{c|}{RS4D (prompt)}&{SSM} & {67.7} & 89.0 & 77.6 & 64.4 & 88.3 & 75.0 &\cellcolor{gray}38.83&\cellcolor{darkgray}11.75&\cellcolor{gray}108.8 \\

	\bottomrule
\end{tabular}
}
\end{table*}

The quantitative comparison results on the WHU dataset are presented in Tab. \ref{tab:Comparisons_other_methods_whu}, with visual examples shown in Fig. \ref{fig:whu2}. While RS4D significantly outperforms traditional models, it performs marginally below certain foundation models. Specifically, RS4D (box) achieves scores of 70.3/70.1 ($\text{AP}_{\text{bbox}}$/$\text{AP}_{\text{mask}}$), which is comparable to the SAM-seg (Mask RCNN) foundation model (70.3/70.1), but falls short of the current state-of-the-art foundation model RSPrompter (query), which achieves 72.5/72.5.

\subsubsection{Comparative Analysis on the NWPU Dataset}

\begin{table*}[!t]  
\caption{Comparison with other methods on the NWPU dataset \cite{NWPU}.}
\label{tab:Comparisons_other_methods_nwpu}
\centering
\resizebox{\linewidth}{!}{
\begin{tabular}{c|c|c c c c c c c c c}
	\toprule
 \multicolumn{1}{c|}{\shortstack{Method}}& $\text{Backbone}$ & $\text{AP}_{\text{bbox}}$ & $\text{AP}^{50}_{\text{bbox}}$ & $\text{AP}^{75}_{\text{bbox}}$ & $\text{AP}_{\text{mask}}$ & $\text{AP}^{50}_{\text{mask}}$ & $\text{AP}^{75}_{\text{mask}}$&Params (M)&$\text{bb-Params (M)}$&$\text{bb-FLOPs (G)}$\\
        \midrule
 \multicolumn{1}{c|}{Mask RCNN \cite{Mask_r-cnn}}&{ResNet50} & 62.3 & 88.3 & 75.2 & 59.7 & 89.2 & 65.6&44.18&\cellcolor{gray}25.56&172.69 \\
 \multicolumn{1}{c|}{MS RCNN \cite{Mask_scoring_r-cnn}}&{ResNet50} & 62.3 & 88.6 & 73.1 & 60.7 & 88.7 & 67.7&60.74&\cellcolor{gray}25.56&172.69 \\
 \multicolumn{1}{c|}{HTC \cite{HTC}} &{ResNet50}& 63.9 & 88.9 & 75.4 & 60.9 & 88.6 & 64.4&46.59&\cellcolor{gray}25.56&172.69 \\
 \multicolumn{1}{c|}{SOLO v2 \cite{solov2}}&{ResNet50} & - & - & - & 50.9 & 77.5 & 54.1&46.58&\cellcolor{gray}25.56&172.69 \\
 \multicolumn{1}{c|}{SCNet \cite{Scnet}}&{ResNet50} & 60.0 & 87.5 & 69.1 & 58.1 & 87.4 & 62.0&94.80&\cellcolor{gray}25.56&172.69 \\
 \multicolumn{1}{c|}{CondInst \cite{Coninst}} &{ResNet50}& 62.3 & 87.8 & 73.3 & 59.0 & 88.5 & 62.8&34.16&\cellcolor{gray}25.56&172.69 \\
 \multicolumn{1}{c|}{BoxInst \cite{boxinst}} &{ResNet50}& 64.8 & 89.3 & 73.0 & 47.6 & 77.2 & 51.3&35.14&\cellcolor{gray}25.56&172.69  \\
 \multicolumn{1}{c|}{Mask2Former \cite{maskformer}}&{ResNet50} & {57.4} & 75.5 & 63.7 & 58.8 & 83.1 & 63.5&44.02&\cellcolor{gray}25.56&172.69 \\
 \multicolumn{1}{c|}{CATNet \cite{CATNet}} &{ResNet50}& {63.2} & 89.0 & 73.8 & 60.4 & 89.6 & 65.5&54.77&\cellcolor{gray}25.56&172.69 \\
 \multicolumn{1}{c|}{HQ-ISNet \cite{HQ-ISNet}} &{ResNet50}& {63.5} & 89.9 & 75.0 & 60.4 & 89.6 & 64.1&80.56&\cellcolor{gray}25.56&172.69 \\
\multicolumn{1}{c|}{Cascade Mask RCNN \cite{cai2019cascade}}&{ResNet50} & {67.5} & \cellcolor{gray}92.4 & 77.3 & 63.5 & \cellcolor{darkgray}92.7 & 66.9&77.02&\cellcolor{gray}25.56&172.69 \\
  \midrule
\multicolumn{1}{c|}{SwinV2(HTC++) \cite{liu2022swin}}&{SwinV2-Huge} & {43.5} &  79.5 & 43.4 & {39.7} & {66.6} & {40.4}&709.09&655.14&303.37 \\
\multicolumn{1}{c|}{CBNetv2 \cite{liang2022cbnet}}&{Swin-Tiny} & {41.0} & 70.3 &  42.4 & {40.8} & {68.8} & {41.5}&75.58&28.30&\cellcolor{darkgray}12.47 \\
\multicolumn{1}{c|}{ViTDet \cite{li2022exploring}}&{Swin-Huge} & {54.8} & 85.4 &  64.7 & {49.6} & {80.4} & {56.6}&679.42&655.58&375.10 \\ 
 \multicolumn{1}{c|}{SAM-seg (Mask RCNN) \cite{RSPrompter}}&{ViT-Base} & \cellcolor{gray}{68.8} & 92.2 & \cellcolor{gray}80.1 & \cellcolor{gray}65.2 & \cellcolor{gray}92.0 & \cellcolor{darkgray}71.6&309.69&89.67&972 \\
 \multicolumn{1}{c|}{SAM-seg (Mask2Former) \cite{RSPrompter}}&{ViT-Base} & {63.1} & 86.3 & 70.6 & 64.3 & 89.6 & 70.1&112.16&89.67&972 \\
 \multicolumn{1}{c|}{SAM-cls \cite{RSPrompter}}&{ViT-Base} & {40.2} & 57.1 & 44.5 & 44.0 & 66.0 & 47.6&{-}&{-}&{-} \\
 \multicolumn{1}{c|}{SAM-det \cite{RSPrompter}}&{ViT-Base} & {64.2} & 89.6 & 74.6 & 51.5 & 74.8 & 54.0&135.08&89.67&972\\
 \multicolumn{1}{c|}{RSPrompter (anchor) \cite{RSPrompter}}&{ViT-Base} & \cellcolor{darkgray}{70.3} & \cellcolor{darkgray}93.6 & \cellcolor{darkgray}81.0 & 66.1 & \cellcolor{darkgray}92.7 & 70.6&117.09&89.67&972 \\
 \multicolumn{1}{c|}{RSPrompter (query) \cite{RSPrompter}}&{ViT-Base} & {68.4} & 90.3 & 74.0 & \cellcolor{darkgray}67.5 & 91.7 & 74.8&101.02&89.67&972 \\
 \midrule
 \multicolumn{1}{c|}{RS4D (box)}&{SSM} &61.7&85.8&70.1&60.8&90.7&64.4&\cellcolor{darkgray}33.20&\cellcolor{darkgray}11.75&\cellcolor{gray}108.8\\
 \multicolumn{1}{c|}{RS4D (prompt)}&{SSM} & {61.5} & 86.9 & 73.7 & 60.3 & 87.4 & 63.3 &\cellcolor{gray}38.83&\cellcolor{darkgray}11.75&\cellcolor{gray}108.8 \\
	\bottomrule
\end{tabular}
}
\end{table*}

The experimental results and visualizations on the NWPU dataset are presented in Tab. \ref{tab:Comparisons_other_methods_nwpu} and Fig. \ref{fig:nwpu1}, respectively. On this dataset, RS4D shows competitive performance with clear efficiency advantages, although its gains over existing methods are more modest than those on SSDD and WHU. Specifically, RS4D (box) achieves 61.7/60.8 ($\text{AP}_{\text{bbox}}$/$\text{AP}_{\text{mask}}$), while RS4D (prompt) achieves 61.5/60.3($\text{AP}_{\text{bbox}}$/$\text{AP}_{\text{mask}}$). These results indicate that the proposed method remains competitive in mask prediction under a much lighter backbone, but it does not outperform the strongest baselines in ($AP_{\text{bbox}}$). This relatively moderate performance can be attributed to the HiPPO-based information compression in Mamba, which may overlook some small objects and densely arranged instances in NWPU, especially under the limited amount of downstream training data.

\subsubsection{Model Efficiency Analysis}

\begin{table}[!t] 
\caption{
A comparison of RS4D's backbone with existing advanced backbone architectures in terms of FLOPs and parameter counts.
}
\label{tab:Comparisons_other_methods_backbone}
\centering
\resizebox{0.9\linewidth}{!}{
\begin{tabular}{l |c c c}
 \toprule
Method & Parameters (M) & FLOPs (G) \\
    \midrule
{ResNet18 \cite{resnet}} & 11.69 & 76.32 \\
{EfficientNet b0 \cite{efficientnet}} & 5.29 & 17.56 \\
{AlexNet \cite{alexnet}} & 61.10 & 30.02 \\
{Vgg16 \cite{vgg16}} & 138.00 & 642.00 \\
{MobileNet v2 \cite{mobilenetv2}} & 3.51 & 13.92 \\
{DenseNet \cite{densenet}} & 28.68 & 330.00 \\
{GoogleNet \cite{googlenet}} & 13.01 & 63.28\\
{ConvNet-b \cite{convnet-b}} & {88.59} & 642.51  \\
{ViT-Base \cite{trandformers}} & {89.67} & 972.00 \\
{RS4D (Ours)} & {11.75} & 108.8 \\
\bottomrule
\end{tabular}
}
\end{table}

\begin{figure}
\centering
\includegraphics[width=1\linewidth]{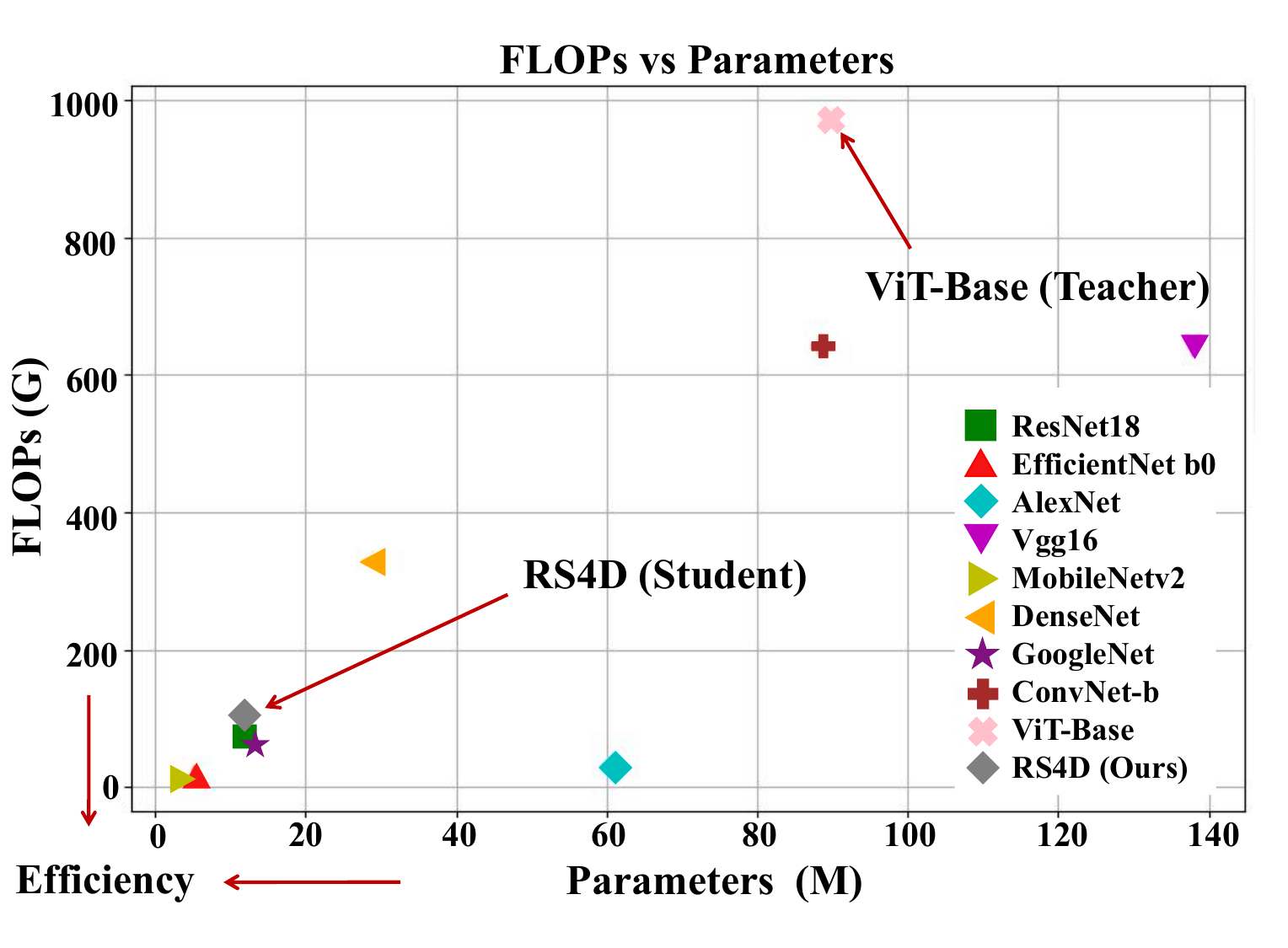}
\vspace{-4ex}
\caption{
A comprehensive comparison of computational efficiency across different models.
}
\label{fig:flops_vs_parameters}
\end{figure}

We first evaluate the efficiency of our backbone in terms of parameter count and computational complexity (FLOPs), comparing it with established architectures including ResNet18 \cite{resnet}, EfficientNet b0 \cite{efficientnet}, AlexNet \cite{alexnet}, VGG16 \cite{vgg16}, MobileNet v2 \cite{mobilenetv2}, DenseNet \cite{densenet}, GoogleNet \cite{googlenet}, ConvNet-b \cite{convnet-b}, and ViT \cite{trandformers}. The comparative analysis is presented in Tab. \ref{tab:Comparisons_other_methods_backbone} and Fig. \ref{fig:flops_vs_parameters}. The backbone of RS4D demonstrates comparable parameter counts and FLOPs to ultra-lightweight CNN models such as MobileNetv2 and EfficientNet b0. Notably, compared to ViT, our model achieves an $\times 8$ reduction in parameter count and a $\times 9$ reduction in FLOPs, making it particularly suitable for real-time applications and resource-constrained environments.

Furthermore, we conducted a comprehensive comparison of model efficiency and instance segmentation performance between RS4D and other instance segmentation models. As shown in Tab. \ref{tab:Comparisons_other_methods_overall} and Fig. \ref{fig:det1}, on the SSDD dataset, RS4D achieves comparable or slightly superior metrics to the large model RSPrompter while requiring only 1/4 of the parameters and 1/9 of the FLOPs. Additionally, when compared to models with similar parameter counts such as CondInst and BoxInst, RS4D demonstrates superior performance.

\begin{table*}[!t]
\caption{
Model efficiency and performance comparisons on the SSDD dataset\cite{SSDD}.
}
\label{tab:Comparisons_other_methods_overall}
\centering
\resizebox{\linewidth}{!}{
\begin{tabular}{c|c c c c c c c c}

 \toprule
Metrics & {Mask RCNN \cite{Mask_r-cnn}} & {MS RCNN \cite{Mask_scoring_r-cnn}} & {HTC \cite{HTC}}& {SCNet \cite{Scnet}} & {CondInst \cite{Coninst}} & {BoxInst \cite{boxinst}} & RSPrompter \cite{RSPrompter}& {RS4D (Ours)} \\
    \midrule
{Parameters (M)} &44.18&60.74&46.59&94.80&34.16&35.14&117.09&33.20 \\
{$\text{AP}_{\text{bbox}}$} & 67.7&67.8&69.3&66.9&68.1&62.8&70.4&70.7 \\
{$\text{AP}_{\text{mask}}$} & 64.3&64.9&64.1&64.9&62.5&45.2&66.8&70.0 \\
	\bottomrule
\end{tabular}
}
\end{table*}

\subsection{Ablation Study}

In this section, we conduct comprehensive ablation studies to validate the effectiveness of our proposed method and its components.

\subsubsection{Effects of Knowledge Distillation}

\begin{figure}
\centering
\includegraphics[width=1\linewidth]{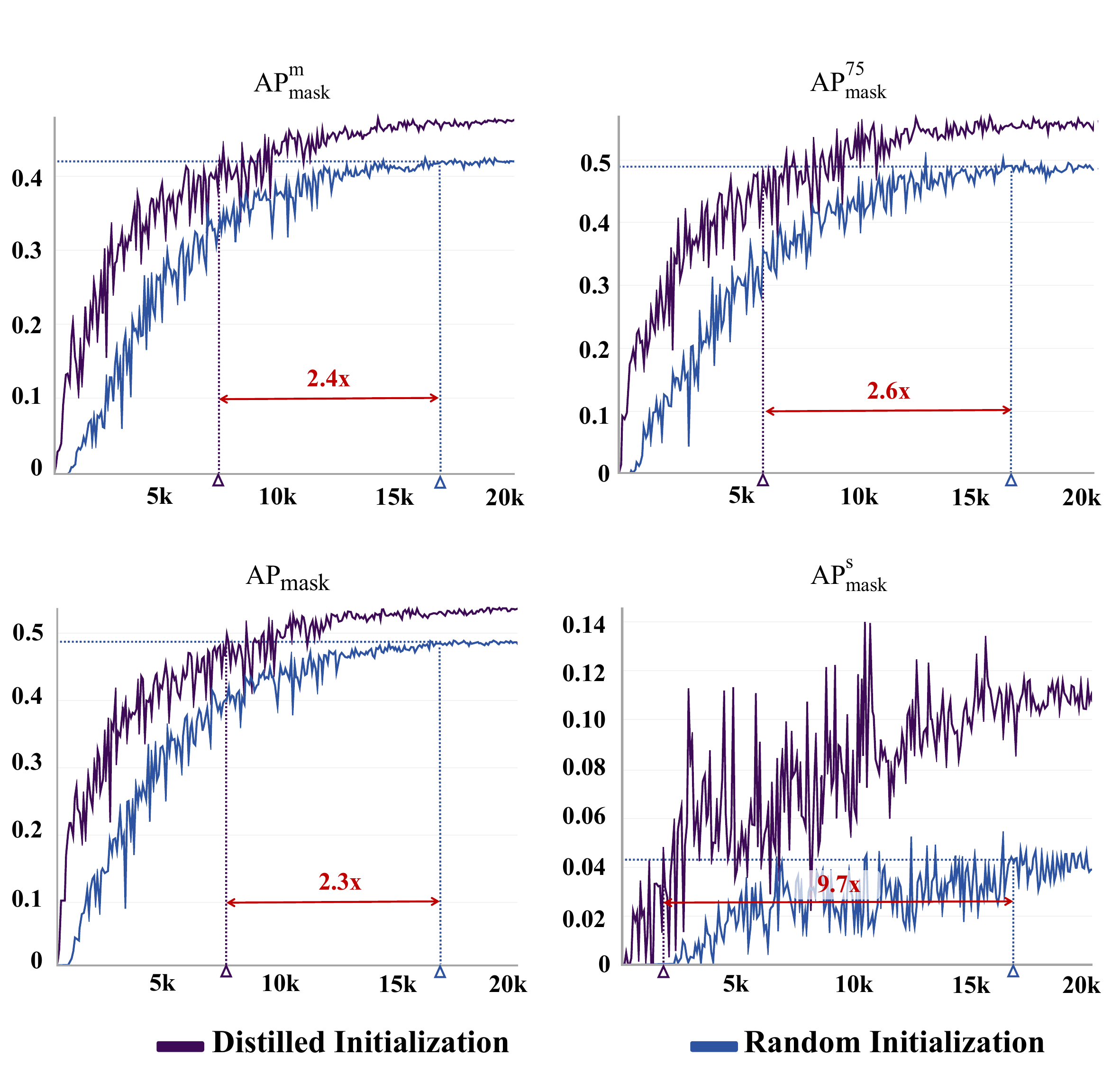}
\vspace{-4ex}
\caption{
Performance across training steps for backbone initialization with and without distilled weights on the NWPU dataset
}
\label{fig:other2}
\end{figure}

To investigate the effect of model initialization, we conducted comparative experiments using distilled weights versus random initialization for the backbone. The experiments were performed on the NWPU dataset, with results presented in Tab. \ref{tab:Comparisons_other_methods_dis} and Fig. \ref{fig:other2}. The results demonstrate that initializing network parameters with distilled weights achieved superior performance compared to random initialization without the foundational model prior knowledge. As shown in Fig. \ref{fig:other2}, the model with distilled weight initialization also exhibited better convergence characteristics.

\begin{table}[!t] 
\caption{
Effect of backbone initialization with and without distilled weights on the NWPU dataset\cite{NWPU}.
}
\label{tab:Comparisons_other_methods_dis}
\centering
\resizebox{0.85\linewidth}{!}{
\begin{tabular}{c|c c c c}
\toprule
 \multicolumn{1}{c|}{\shortstack{Dist. Init.}} &  $\text{AP}_{\text{mask}}$ &  $\text{AP}^{75}_{\text{mask}}$ &  $\text{AP}^{ \text{m}}_{\text{mask}}$ &  $\text{AP}^{\text{s}}_{\text{mask}}$\\
    \midrule
\multicolumn{1}{c|}{w/} &  {52.2} &  {54.5} &  {46.5} &  {9.9}\\
\multicolumn{1}{c|}{w/o} &  {48.3} &  {49.0} &  {41.6} &  {5.3}\\
 \bottomrule
\end{tabular}
}
\end{table}

\subsubsection{Effects of Feature Quantity on Neck Fusion}

This section investigates the impact of neck feature composition on instance segmentation performance. We constructed the neck component by selecting features from different hidden layers of the SSM backbone. The experimental results comparing different layer combinations are presented in Tab. \ref{tab:Comparisons_other_methods_ab3}, Tab. \ref{tab:Comparisons_other_methods_ab3_whu}, and Tab. \ref{tab:Comparisons_other_methods_ab3_nwpu}. They were conducted on the SSDD, WHU, and NWPU datasets, respectively. The results demonstrate that appropriately increasing the number of selected hidden layers facilitates the fusion of deep and shallow features, thereby enhancing semantic segmentation representation. When using only a single layer (\textit{i.e.}, the final layer features), the model exhibited the poorest performance.

\begin{table}[!t]  
\caption{Performance of neck fusion's different multi-layer feature selection on the SSDD dataset\cite{SSDD}.}
\label{tab:Comparisons_other_methods_ab3}
\centering
\resizebox{\linewidth}{!}{
\begin{tabular}{c|c c c c c c}
	\toprule
 \multicolumn{1}{c|}{Layers} &$\text{AP}_{\text{bbox}}$ & $\text{AP}^{50}_{\text{bbox}}$ & $\text{AP}^{75}_{\text{bbox}}$ & $\text{AP}_{\text{mask}}$ & $\text{AP}^{50}_{\text{mask}}$ & $\text{AP}^{75}_{\text{mask}}$\\
     \midrule
\multicolumn{1}{c|}{24} &{70.5} & {97.5} & {85.2} & {68.7} & {97.4} & {87.2}\\
\multicolumn{1}{c|}{12}& 70.7&97.4&86.4&70.0&97.4&87.5\\
\multicolumn{1}{c|}{6} &{69.3} & {96.7} & {84.4} & {69.3} & {95.8} & {86.7}\\
\multicolumn{1}{c|}{4} &71.0&97.7&86.5&70.0&97.6&86.8\\
\multicolumn{1}{c|}{3} &69.3&97.6&84.1&69.6&97.6&86.3\\
\multicolumn{1}{c|}{2} &69.6&97.7&85.5&69.0&96.7&85.6\\
\multicolumn{1}{c|}{1} &{1.70} & {0.67} & {0.01} & {0.06} & {0.28} & {0.01}\\
\bottomrule
\end{tabular}
}
\end{table}

\begin{table}[!t]  
\caption{Performance of neck fusion's different multi-layer feature selection on the WHU dataset\cite{ji2018fully}.}
\label{tab:Comparisons_other_methods_ab3_whu}
\centering
\resizebox{\linewidth}{!}{
\begin{tabular}{c|c c c c c c}
	\toprule
 \multicolumn{1}{c|}{Layers} &$\text{AP}_{\text{bbox}}$ & $\text{AP}^{50}_{\text{bbox}}$ & $\text{AP}^{75}_{\text{bbox}}$ & $\text{AP}_{\text{mask}}$ & $\text{AP}^{50}_{\text{mask}}$ & $\text{AP}^{75}_{\text{mask}}$\\
     \midrule
\multicolumn{1}{c|}{24} &68.4&89.6&77.7&68.1&89.8&78.8\\
\multicolumn{1}{c|}{12}&70.3&89.3&79.2&70.1&89.5&80.3\\
\multicolumn{1}{c|}{6} &68.8&89.8&78.7&69.0&89.9&79.8\\
\multicolumn{1}{c|}{4} &69.1&89.9&78.9&69.4&90.1&80.0\\
\multicolumn{1}{c|}{3} &68.7&89.7&78.6&68.9&89.9&79.8\\
\multicolumn{1}{c|}{2} &68.6&89.7&78.5&69.3&89.8&79.9\\
\multicolumn{1}{c|}{1} &6.3&21.7&1.0&1.8&10.4&0.1\\
\bottomrule
\end{tabular}
}
\end{table}

\begin{table}[!t]  
\caption{Performance of neck fusion's different multi-layer feature selection on the NWPU \cite{NWPU}.}
\label{tab:Comparisons_other_methods_ab3_nwpu}
\centering
\resizebox{\linewidth}{!}{
\begin{tabular}{c|c c c c c c}
	\toprule
 \multicolumn{1}{c|}{Layers} &$\text{AP}_{\text{bbox}}$ & $\text{AP}^{50}_{\text{bbox}}$ & $\text{AP}^{75}_{\text{bbox}}$ & $\text{AP}_{\text{mask}}$ & $\text{AP}^{50}_{\text{mask}}$ & $\text{AP}^{75}_{\text{mask}}$\\
     \midrule
\multicolumn{1}{c|}{24} &61.2&86.6& 72.3& 59.0& 85.6& 63.4\\
\multicolumn{1}{c|}{12}&60.6& 85.0& 69.3& 57.7& 83.6& 62.1 \\
\multicolumn{1}{c|}{6} &58.4& 82.6& 67.0& 56.1& 82.5& 59.8\\
\multicolumn{1}{c|}{4} &60.8& 87.8& 69.3& 58.2& 86.7& 62.4\\
\multicolumn{1}{c|}{3} &59.9&84.8&70.7&57.1&84.2&61.6\\
\multicolumn{1}{c|}{2} &59.7& 85.3& 69.0& 57.7& 83.6& 60.7\\
\multicolumn{1}{c|}{1} &3.8 &10.7 &1.9 &2.2& 7.5 &0.5 \\
\bottomrule
\end{tabular}
}
\end{table}

\subsubsection{Effects of Noising and Masking in Distillation}

Tab. \ref{tab:Comparisons_other_methods_ab4} demonstrates the impact of varying noise levels and masking proportions against the accuracy using the SSDD dataset. Tab. \ref{tab:Comparisons_other_methods_ab4_whu} and Tab. \ref{tab:Comparisons_other_methods_ab4_nwpu} were conducted on the WHU and NWPU datasets respectively. The analysis reveals that controlled noise injection enhances object detection capabilities, while systematic masking strategies improve segmentation accuracy. Optimal performance was achieved with a 20\% perturbation ratio; higher ratios led to performance degradation. The strategic application of noise and masking techniques during training resulted in enhanced model robustness and improved overall performance metrics.

\begin{table}[!t]
\caption{
Instance segmentation performance of varying noise and masking ratios in knowledge distillation on the SSDD dataset\cite{SSDD}.
}
\label{tab:Comparisons_other_methods_ab4}
\centering
\resizebox{\linewidth}{!}{
\begin{tabular}{c|c|c c c c c c}
	\toprule
 \multicolumn{1}{c|}{\shortstack{Noise}}&{Mask} &$\text{AP}_{\text{bbox}}$ & $\text{AP}^{50}_{\text{bbox}}$ & $\text{AP}^{75}_{\text{bbox}}$ & $\text{AP}_{\text{mask}}$ & $\text{AP}^{50}_{\text{mask}}$ & $\text{AP}^{75}_{\text{mask}}$\\
     \midrule
\multicolumn{1}{c|}{0\%}&{0\%} &70.4&96.8&86.2&68.7&95.7&85.9\\
\multicolumn{1}{c|}{10\%}&{0\%} &70.8&98.4&87.4&69.0&97.5&84.7\\
\multicolumn{1}{c|}{0\%}&{10\%} &70.2&97.7&85.3&69.3&97.7&87.5\\
\multicolumn{1}{c|}{10\%}&{10\%} &{69.2} & {96.6} & {83.6} & {69.3} & {97.4} & {86.7}\\
\multicolumn{1}{c|}{20\%}&{20\%}&70.7&97.4&86.4&70.0&97.4&87.5\\ 
\multicolumn{1}{c|}{30\%}&{30\%}&69.2&97.6&83.5&69.2&96.8&86.7 \\
\bottomrule
\end{tabular}
}
\end{table}

\begin{table}[!t]
\caption{
Instance segmentation performance of varying noise and masking ratios in knowledge distillation on the WHU dataset\cite{ji2018fully}.
}
\label{tab:Comparisons_other_methods_ab4_whu}
\centering
\resizebox{\linewidth}{!}{
\begin{tabular}{c|c|c c c c c c}
	\toprule
 \multicolumn{1}{c|}{\shortstack{Noise}}&{Mask} &$\text{AP}_{\text{bbox}}$ & $\text{AP}^{50}_{\text{bbox}}$ & $\text{AP}^{75}_{\text{bbox}}$ & $\text{AP}_{\text{mask}}$ & $\text{AP}^{50}_{\text{mask}}$ & $\text{AP}^{75}_{\text{mask}}$\\
     \midrule
\multicolumn{1}{c|}{0\%}&{0\%} &69.0&88.4&78.7&69.4&89.3&80.0\\
\multicolumn{1}{c|}{10\%}&{0\%} &69.1&89.8 &78.7 &68.5 &89.9 &79.6\\
\multicolumn{1}{c|}{0\%}&{10\%} &68.8&89.2 &78.8 &69.0 &89.4 &80.1\\
\multicolumn{1}{c|}{10\%}&{10\%} &{69.0} & {89.7} & {78.7} & {69.2} & {89.9} & {79.8}\\
\multicolumn{1}{c|}{20\%}&{20\%}&70.3&89.3&79.2&70.1&89.5&80.3\\ 
\multicolumn{1}{c|}{30\%}&{30\%}&68.0&89.2&77.9&68.2&89.3&79.1 \\
\bottomrule
\end{tabular}
}
\end{table}

\begin{table}[!t]
\caption{
Instance segmentation performance of varying noise and masking ratios in knowledge distillation on the NWPU dataset\cite{NWPU}.
}
\label{tab:Comparisons_other_methods_ab4_nwpu}
\centering
\resizebox{\linewidth}{!}{
\begin{tabular}{c|c|c c c c c c}
	\toprule
 \multicolumn{1}{c|}{\shortstack{Noise}}&{Mask} &$\text{AP}_{\text{bbox}}$ & $\text{AP}^{50}_{\text{bbox}}$ & $\text{AP}^{75}_{\text{bbox}}$ & $\text{AP}_{\text{mask}}$ & $\text{AP}^{50}_{\text{mask}}$ & $\text{AP}^{75}_{\text{mask}}$\\
     \midrule
\multicolumn{1}{c|}{0\%}&{0\%} &57.0&83.8&66.2&54.7&81.6&57.8\\
\multicolumn{1}{c|}{10\%}&{0\%} &56.2&80.2&65.3&54.3&79.2&57.4\\
\multicolumn{1}{c|}{0\%}&{10\%} &58.6&83.4&67.8&56.4&82.6&60.4\\
\multicolumn{1}{c|}{10\%}&{10\%} &{59.2} & {84.6} & {67.7} & {56.5} & {84.0} & {60.0}\\
\multicolumn{1}{c|}{20\%}& {20\%}& 60.6&85.0&69.3&57.7&83.6&62.1\\ 
\multicolumn{1}{c|}{30\%}&{30\%}&57.6&81.9&64.6&54.8&81.8&57.1 \\
\bottomrule
\end{tabular}
}
\end{table}

\subsubsection{Effects of Backbone Depth and Width}

Our parametric study using the SSDD dataset examined how varying the depth (number of layers) and width (dimension) of the SSM backbone affected model performance. As illustrated in Tab. \ref{tab:Comparisons_other_methods_ab5}, Tab. \ref{tab:Comparisons_other_methods_ab5_whu}, and Tab. \ref{tab:Comparisons_other_methods_ab5_nwpu}, modest increases in both parameters yielded improved overall performance. However, arbitrarily increasing model parameters without considering computational resources and dataset constraints proves counterproductive, as it substantially elevates the risk of overfitting.

\subsubsection{Effects of Distillation Data Volume}

Our experimental study examined the impact of distillation data volume on model performance, starting with only 0.1\% of the SA-1B dataset\cite{SAM}. We gradually increased the data volume to 10k (0.1\%), 25k (0.25\%), 100k (1\%), and 1000k (10\%) samples. As shown in Tab. \ref{tab:Distillation_Data_Volume}, increasing the distillation data volume consistently improves downstream RS instance segmentation performance, especially $\text{AP}_{\text{mask}}$, indicating that the proposed distillation procedure can benefit from larger-scale unlabeled data when computational resources permit. This suggests that a larger dataset provides richer information, enabling better generalization. Conversely, the suboptimal performance observed in the NWPU dataset highlights the importance of sufficient training data for effective knowledge transfer. These results suggest that the required distillation volume depends on the downstream data regime. Smaller and more category-diverse datasets, such as NWPU, can benefit more from larger distilled subsets, whereas a small subset can already provide an effective and efficient initialization for datasets such as SSDD and WHU. Therefore, in this work, the choice of distilled data volume should be understood as an empirical efficiency-performance trade-off rather than a claim of a universal saturation threshold.

\subsubsection{Effects of Frozen-backbone}

To assess representation quality, we additionally evaluate different backbone settings. Among them, "all-params" and "frozen-backbone" use the pre-trained parameters we distilled, with the former being fully fine-tuned and the latter only fine-tuning the neck and head. "Zero-init" randomly initializes the parameters and performs full parameter fine-tuning. "imagenet-init" replaces the backbone with the ImageNet pre-trained EfficientNet-B1 and also performs full parameter fine-tuning. The frozen protocol still achieves strong performance, as seen in the added row in Tab. \ref{tab:frozen-backbone}, indicating that the distilled backbone learns transferable representations rather than relying solely on end-to-end adaptation. At the same time, we also compared RS4D with EfficientNet-B1, which has a similar number of backbone parameters and is pretrained on ImageNet. In the scenario of full parameter fine-tuning, RS4D demonstrates a significant advantage.

\begin{table}[!t] 
\caption{
Instance segmentation performance of SSM backbone's depth and width on the SSDD dataset\cite{SSDD}.
}
\label{tab:Comparisons_other_methods_ab5}
\centering
\resizebox{\linewidth}{!}{
\begin{tabular}{c|c|c c c c c c}
\toprule
\multicolumn{1}{c|}{\shortstack{Depth}}&{Width} &$\text{AP}_{\text{bbox}}$ & $\text{AP}^{50}_{\text{bbox}}$ & $\text{AP}^{75}_{\text{bbox}}$ & $\text{AP}_{\text{mask}}$ & $\text{AP}^{50}_{\text{mask}}$ & $\text{AP}^{75}_{\text{mask}}$\\
\midrule
\multicolumn{1}{c|}{16}&{256}&{68.1}&{97.3}&{81.7}&{68.7}&{97.4}&{84.4}\\
\multicolumn{1}{c|}{24}&{256}&70.7&97.4&86.4&70.0&97.4&87.5\\
\multicolumn{1}{c|}{24}&{320} &70.5&98.5&86.9&69.8&98.5&86.9\\
\multicolumn{1}{c|}{28}&{256}&69.7&96.6&85.1&69.0&96.6&87.1\\
\multicolumn{1}{c|}{48}&{256} &69.8&97.5&85.5&68.2&97.5&85.7\\
\multicolumn{1}{c|}{48}&{320}&69.4&97.4&84.5&69.8&97.5&87.0\\
\bottomrule
\end{tabular}
}
\end{table}

\begin{table}[!t] 
\caption{
Instance segmentation performance of SSM backbone's depth and width on the WHU dataset\cite{ji2018fully}.
}
\label{tab:Comparisons_other_methods_ab5_whu}
\centering
\resizebox{\linewidth}{!}{
\begin{tabular}{c|c|c c c c c c}
\toprule
\multicolumn{1}{c|}{\shortstack{Depth}}&{Width} &$\text{AP}_{\text{bbox}}$ & $\text{AP}^{50}_{\text{bbox}}$ & $\text{AP}^{75}_{\text{bbox}}$ & $\text{AP}_{\text{mask}}$ & $\text{AP}^{50}_{\text{mask}}$ & $\text{AP}^{75}_{\text{mask}}$\\
\midrule
\multicolumn{1}{c|}{16}&{256}&57.9&84.0&66.2&56.5&84.2&65.4 \\
\multicolumn{1}{c|}{24}&{256}&70.3&89.3&79.2&70.1&89.5&80.3\\
\multicolumn{1}{c|}{24}&{320} &69.1&88.3&78.7&68.9&89.3&79.2\\
\multicolumn{1}{c|}{28}&{256}&68.3&88.3&77.8&68.3&88.5&79.0\\
\multicolumn{1}{c|}{48}&{256} &66.6&88.8&76.6&65.2&89.0&76.6\\
\bottomrule
\end{tabular}
}
\end{table}

\begin{table}[!t] 
\caption{
Instance segmentation performance of SSM backbone's depth and width on the NWPU dataset\cite{NWPU}.
}
\label{tab:Comparisons_other_methods_ab5_nwpu}
\centering
\resizebox{\linewidth}{!}{
\begin{tabular}{c|c|c c c c c c}
\toprule
\multicolumn{1}{c|}{\shortstack{Depth}}&{Width} &$\text{AP}_{\text{bbox}}$ & $\text{AP}^{50}_{\text{bbox}}$ & $\text{AP}^{75}_{\text{bbox}}$ & $\text{AP}_{\text{mask}}$ & $\text{AP}^{50}_{\text{mask}}$ & $\text{AP}^{75}_{\text{mask}}$\\
\midrule
\multicolumn{1}{c|}{16}&{256}&{52.1}&{7.0}&{58.2}&{49.2}&{74.6}&{51.7}\\
\multicolumn{1}{c|}{24}&{256}&60.6&85.0&69.3&57.7&83.6&62.1\\
\multicolumn{1}{c|}{24}&{320} & 58.9&81.8&68.1&58.0&82.0&63.0\\
\multicolumn{1}{c|}{28}&{256}& 56.2&83.5&65.6&53.9&82.2&58.6\\
\multicolumn{1}{c|}{48}&{256} &52.9&78.8&59.1&51.2&77.3&53.1\\
\bottomrule
\end{tabular}
}
\end{table}

\begin{table}[!t]  
\caption{Impact of different data volume in distillation on the NWPU dataset\cite{NWPU}.}
\label{tab:Distillation_Data_Volume}
\centering
\resizebox{\linewidth}{!}{
\begin{tabular}{c|c c c c c c}
	\toprule
 \multicolumn{1}{c|}{Data} &$\text{AP}_{\text{bbox}}$ & $\text{AP}^{50}_{\text{bbox}}$ & $\text{AP}^{75}_{\text{bbox}}$ & $\text{AP}_{\text{mask}}$ & $\text{AP}^{50}_{\text{mask}}$ & $\text{AP}^{75}_{\text{mask}}$\\
     \midrule
\multicolumn{1}{c|}{10k (0.1\%)} &{56.9} & 82.4 & 65.6 & 54.8 & 81.5 & 55.5 \\
\multicolumn{1}{c|}{25k (0.25\%)} &{57.9} & {81.6} & {67.4} & {55.1} & {80.7} & {57.2}\\
\multicolumn{1}{c|}{100k (1\%)} &{60.6} & {85.0} & {69.3} & {57.7} & {83.6} & {62.1}\\
\multicolumn{1}{c|}{1000k (10\%)} &61.7&85.8 &70.1& 60.8& 90.7& 64.4\\
\bottomrule
\end{tabular}
}
\end{table}

\begin{table}[!t]  
\caption{Impact of frozen-backbone in distillation on the SSDD dataset\cite{SSDD}.}
\label{tab:frozen-backbone}
\centering
\resizebox{\linewidth}{!}{
\begin{tabular}{c|c c c c c c}
	\toprule
 \multicolumn{1}{c|}{Setting} &$\text{AP}_{\text{bbox}}$ & $\text{AP}^{50}_{\text{bbox}}$ & $\text{AP}^{75}_{\text{bbox}}$ & $\text{AP}_{\text{mask}}$ & $\text{AP}^{50}_{\text{mask}}$ & $\text{AP}^{75}_{\text{mask}}$\\
     \midrule
\multicolumn{1}{c|}{all-params} &70.7&97.4&86.4&70.0&97.4&87.5 \\
\multicolumn{1}{c|}{frozen-backbone} &68.9&96.6&84.4&68.6&96.6&85.7\\
\multicolumn{1}{c|}{Zero-init} &66.3 &96.2& 82.3& 66.0& 96.7& 82.0\\
\multicolumn{1}{c|}{imagenet-init} &68.6&98.2&83.6&69.4&97.8&86.1\\
\bottomrule
\end{tabular}
}
\end{table}

\subsection{Discussion and Limitation}

As a lightweight instance segmentation architecture, RS4D shows considerable promise for model compression and edge computing applications. Through the transformation of sparse attention space into dense state space, the model effectively distills essential information while achieving significant compression in the state space domain. This approach offers a particularly promising solution for foundation models with billions of parameters, potentially enabling efficient knowledge storage and rapid inference capabilities.

RS4D demonstrates superior performance compared to both conventional approaches and foundation models when tested on WHU and SSDD datasets. However, its performance on the NWPU dataset only exceeds that of some traditional models. This limitation can be attributed to the visual information compression process utilizing Hippo matrices \cite{hippo}, where small objects that occupy minimal image patches are frequently lost during information compression. While the target size in SSDD is similarly small, the issue is not as significant. However, we observe that this problem becomes more pronounced when the number of training images is limited, as seen in the NWPU dataset. Additionally, the NWPU dataset contains a dense arrangement of numerous objects within the same category, making it easy for many adjacent and semantically similar patches to be overlooked during HiPPO’s information compression. Future enhancements could address this limitation through increased embedding dimensionality, improved local feature integration, and optimized selective mechanism design.

\section{Conclusion}

This paper introduces RS4D, a novel framework for instance segmentation of remote sensing images that leverages State Space Models (SSMs) with linear computational complexity. The framework effectively condenses information from extensive self-attention spaces into compact linear state spaces. Our investigation encompasses three distinct SSM backbone architectures and two implementations for instance segmentation heads. Comprehensive experimental results demonstrate RS4D's exceptional performance in remote sensing image instance segmentation, achieving a $\times 9$ reduction in FLOPs compared to ViT-based methods and a $\times 6$ reduction relative to ConvNet approaches, while showing especially strong performance on SSDD and WHU and competitive efficiency results on NWPU. Notably, RS4D achieves performance on par with the foundation model-based methods while utilizing only 1/4 of the model parameters, making it an efficient solution for deploying lightweight foundation models.

\ifCLASSOPTIONcaptionsoff
  \newpage
\fi

\bibliographystyle{IEEEtran}

\bibliography{references.bib}

\end{CJK*}
\end{document}